\documentclass[sigconf,screen]{acmart}

\AtBeginDocument{%
  \providecommand\BibTeX{{%
    \normalfont B\kern-0.5em{\scshape i\kern-0.25em b}\kern-0.8em\TeX}}}

\copyrightyear{2022}
\acmYear{2022}
\setcopyright{rightsretained}
\acmConference[WWW '22]{Proceedings of the ACM Web Conference 2022}{April 25--29, 2022}{Virtual Event, Lyon, France}
\acmBooktitle{Proceedings of the ACM Web Conference 2022 (WWW '22), April 25--29, 2022, Virtual Event, Lyon, France}
\acmDOI{10.1145/3485447.3512189}
\acmISBN{978-1-4503-9096-5/22/04}

\usepackage{amsmath,amsfonts,bm}

\def\eqref#1{Eq~(\ref{#1})}

\def\1{\bm{1}}

\def\rs{{\textnormal{s}}}

\def\va{{\bm{a}}}

\def\vz{{\bm{z}}}

\DeclareMathAlphabet{\mathsfit}{\encodingdefault}{\sfdefault}{m}{sl}
\SetMathAlphabet{\mathsfit}{bold}{\encodingdefault}{\sfdefault}{bx}{n}

\def\gA{{\mathcal{A}}}

\def\gE{{\mathcal{E}}}

\def\gG{{\mathcal{G}}}

\def\gL{{\mathcal{L}}}
\def\gM{{\mathcal{M}}}

\def\gS{{\mathcal{S}}}

\def\gV{{\mathcal{V}}}

\def\sI{{\mathbb{I}}}

\def\sR{{\mathbb{R}}}

\newcommand{\E}{\mathbb{E}}

\usepackage{graphicx}
\usepackage{amsmath}
\usepackage{cleveref}
\usepackage{mathtools}
\usepackage{xspace}
\usepackage{multicol}
\usepackage{multirow}
\usepackage{booktabs}
\usepackage{caption}
\usepackage{subcaption}

\newcommand{\UGE}{UGE\xspace}

\newcommand{\norm}[1]{\left\lVert#1\right\rVert}
\def \btheta {\bm \theta}

\makeatletter
\def\blfootnote{\gdef\@thefnmark{}\@footnotetext}
\makeatother

\begin{document}
\title[Unbiased Graph Embedding with Biased Graph Observations]{Unbiased Graph Embedding with Biased Graph Observations}

\author{Nan Wang, Lu Lin, Jundong Li, Hongning Wang}
\email{{nw6a, ll5fy, jl6qk, hw5x}@virginia.edu}
\affiliation{%
  \institution{University of Virginia}
  \city{Charlottesville}
  \state{VA}
  \country{USA}
}


\renewcommand{\shortauthors}{Wang, et al.}

\begin{abstract}
Graph embedding techniques are pivotal in real-world machine learning tasks that operate on graph-structured data, such as social recommendation and protein structure modeling. Embeddings are mostly performed on the node level for learning representations of each node. Since the formation of a graph is inevitably affected by certain \emph{sensitive node attributes}, 
the node embeddings can inherit such sensitive information and introduce undesirable biases in downstream tasks. Most existing works impose ad-hoc constraints on the node embeddings to restrict their distributions for unbiasedness/fairness, which however compromise the utility of the resulting embeddings. 
In this paper, we propose a principled new way for unbiased graph embedding by learning node embeddings from an \emph{underlying bias-free graph}, which is not influenced by sensitive node attributes. Motivated by this new perspective, we propose two complementary methods for uncovering such an underlying graph, with the goal of introducing minimum impact on the utility of the embeddings. Both our theoretical justification and extensive experimental comparisons against state-of-the-art solutions demonstrate the effectiveness of our proposed methods. 
\end{abstract}



\maketitle
\footnotetext{Equal contribution between the first two authors.}
\section{Introduction}
\label{sec:intro}
Graph embedding is an indispensable building block in modern machine learning approaches that operate on graph-structured data \cite{perozzi2014deepwalk, tang2015line, grover2016node2vec, kipf2017semisupervised, Goyal2018graph}. Graph embedding methods map each node to a low-dimensional embedding vector that reflects the nodes' structural information from the observed connections in the given graph. These node embeddings are then employed to solve downstream tasks, such as friend recommendation in social networks (i.e., link prediction) or user interest prediction in e-commerce platforms (i.e., node classification) \cite{wang2016structural,ou2016asymmetric}. 

However, the observed node connections in a graph are inevitably affected by certain \emph{sensitive node attributes} (e.g., gender, age, race, religion, etc., of users) \cite{pfeiffer2014attributed}, which are intended to be withheld from many high-stake real-world applications. Without proper intervention, the learned node embeddings can inherit undesired sensitive information and lead to severe bias and fairness concerns in downstream tasks \cite{rahman2019fairwalk,bose2019compositional}.
For example, in social network recommendation, if the users with the same gender are observed to connect more often, the learned embeddings can record such information and lead to gender bias by only recommending friends to a user with the same gender identity. 
Biased node embeddings, when applied in applications such as 
loan application \cite{kulik2000demographics} or criminal justice \cite{berk2021fairness}, may unintentionally favor or disregard one demographic group, causing unfair treatments. 
Besides, from the data privacy perspective, this also opens up the possibility for extraction attacks from the node embeddings \cite{sun2018adversarial}. These realistic and ethical concerns set a higher bar for the graph embedding methods to learn both effective and unbiased embeddings.

There is rich literature in enforcing unbiasedness/fairness in algorithmic decision making, especially in classical classification problems \cite{kamishima2012fairness,zemel2013learning,chouldechova2016fair}. Unbiased graph embedding has just started to attract research attentions in recent years. To date, the most popular recipe for unbiased graph embedding is to add adversarial regularizations to the loss function, such that the sensitive attributes cannot be predicted from the learned embeddings \cite{madras2018learning,bose2019compositional,agarwal2021towards, dai2021say}. 
For example, making a discriminator built on the node embeddings fail to predict the sensitive attributes of the nodes. 
However, such a regularization is only \emph{a necessary condition} for debiasing node embeddings, and it usually hurts the utility of the embeddings (a trivial satisfying solution is to randomize the embeddings). Besides these regularization-based solutions, Fairwalk \cite{rahman2019fairwalk} modifies the random walk strategy in the node2vec algorithm \cite{grover2016node2vec} into two levels: when choosing the next node on a path, it first randomly selects a group defined by sensitive attributes, and then randomly samples a reachable node from that group. DeBayes \cite{buyl2020debayes} proposes to capture the sensitive information by a prior function in Conditional Network Embedding \cite{kang2018conditional}, such that the learned embeddings will not carry the sensitive information. Nevertheless, both Fairwalk and DeBayes are based on specific graph embedding methods; and how to generalize them to other types of graph embedding methods such as GAT \cite{velickovic2018graph} or SGC \cite{wu2019simplifying} is not obvious. 


Moving beyond the existing unbiased graph embedding paradigm, in this paper, we propose a principled new framework for the purpose with theoretical justifications. Our solution is to learn node embeddings from an \emph{underlying bias-free graph} whose edges are generated without influence from sensitive attributes. Specifically, as suggested by \citet{pfeiffer2014attributed}, the generation of a graph can be treated as a two-phase procedure. In the first phase, the nodes are connected with each other solely based on global graph \emph{structural properties}, such as degree distributions, diameter, edge connectivity, clustering coefficients and etc., resulting in an \emph{underlying structural graph}, free of influences from node attributes. In the second phase, the connections are \emph{re-routed} by the node attributes (including both sensitive and non-sensitive attributes). For example, in a social network, users in the same age group tend to be more connected than those in different age groups, leading to the final observed graph biased by the age attribute. Hence, our debiasing principle is to filter out the influence from sensitive attributes on the underlying structural graph to create a bias-free graph (that only has non-sensitive or no attributes) from the observed graph, and then perform embedding learning on the bias-free graph. 


We propose two alternative ways to uncover the bias-free graph from the given graph for learning node embeddings. The first is a weighting-based method, which reweighs the graph reconstruction based loss function with importance sampling on each edge, such that the derived loss is as calculated on the bias-free graph, in expectation. This forms \emph{a sufficient condition} for learning unbiased node embeddings: when the reconstruction loss is indeed defined on the corresponding bias-free graph, the resulting node embeddings are unbiased, since the bias-free graph is independent from the sensitive attributes.  
The second way is via regularization, in which we require that, with and without the sensitive attributes, 
the probabilities of generating an edge between two nodes from their embeddings are the same.
In contrast, this forms \emph{a necessary condition}: when the learning happens on the bias-free graph, the resulting embeddings should not differentiate if any sensitive attributes participated in the generation of observed graph, i.e., the predicted edge generation should be independent from the sensitive attributes. 
These two methods are complementary and can be combined to control the trade-off between utility and unbiasedness. 


Comprehensive experiments on three datasets and several backbone graph embedding models prove the effectiveness of our proposed framework. It achieves encouraging trade-off between unbiasedness and utility of the learned embeddings. Results also suggest that the embeddings from our methods can lead to \emph{fair} predictions in the downstream applications.   
In \Cref{sec:related}, we discuss the related work. We introduce the notation and preliminary knowledge on unbiased graph embedding in \Cref{sec:preliminary}. We formally define the underlying bias-free graph in \Cref{sec:under-graph}, and propose the unbiased graph embedding methods in \Cref{sec:UGE}. We evaluate the proposed methods in \Cref{sec:exp} and conclude in \Cref{sec:conclusion}.

\section{Related Work}
\label{sec:related}
Graph embedding aims to map graph nodes to low-dimensional vector representations such that the original graph can be reconstructed from these node embeddings. Traditional approaches include matrix factorization and spectral clustering techniques \cite{ng2002spectral, belkin2001laplacian}. 
Recent years have witnessed numerous successful advances in deep neural architectures for learning node embeddings.
Deepwalk \cite{perozzi2014deepwalk} and node2vec \cite{grover2016node2vec} utilize a skip-gram \cite{mikolov2013efficient} based objective to recover the node context in random walks on a graph. 
Graph Convolutional Networks (GCNs) learn a node's embedding by aggregating the features from its neighbors supervised by node/edge labels in an end-to-end manner.
These techniques are widely applied in friend or content recommendation \cite{liu2019real,xie2016learning}, protein structure prediction \cite{jumper2021highly}, and many more.


Recent efforts on unbiased and fair graph embedding mainly focus on \textit{pre-processing}, \textit{algorithmic} and \textit{post-processing} steps in the learning pipeline. 
The pre-processing solutions modify the training data to reduce the leakage of sensitive attributes \cite{calmon2017optimized}.
Fairwalk \cite{rahman2019fairwalk} is a typical pre-processing method which modifies the sampling process of random walk on graphs by giving each group of neighboring nodes an equal chance to be chosen. However, such pre-processing may well shift the data distribution and leads the trained model to inferior accuracy and fairness measures.
The \textit{post-processing} methods employ discriminators to correct the learned embeddings to satisfy specific fairness constraints \cite{hardt2016equality}.
However, such ad-hoc post-correction is detached from model training which can heavily degrade model's prediction quality.

Our work falls into the category of \textit{algorithmic} methods, which modify the learning objective to prevent bias from the node embeddings. The most popular algorithmic solution is adding (adversarial) regularizations as constraints to filter out sensitive information \cite{bose2019compositional, dai2020learning, agarwal2021towards}.
Compositional fairness constraints \cite{bose2019compositional} are realized by a composition of discriminators for a set of sensitive attributes jointly trained with the graph embedding model. Similarly, FairGNN \cite{dai2020learning} adopts a fair discriminator but focuses on debiasing with missing sensitive attribute values. Different from regularization based methods. DeBayes \cite{buyl2020debayes} reformulates the maximum likelihood estimation with a biased prior which absorbs the information about sensitive attributes; but this solution is heavily coupled with the specific embedding method thus is hard to generalize.
Our method differs from these previous works by learning embeddings from an underlying bias-free graph. We investigate the generation of the given graph and remove the influence from sensitive attributes in the generative process to uncover a bias-free graph for graph embedding. 

\textit{Generative graph models} \cite{airoldi2008mixed, pfeiffer2014attributed} focus on the statistical process of graph generation by modeling the joint distributions of edges conditioned on node attributes and graph structure.
For instance, Attributed Graph Model (AGM) \cite{pfeiffer2014attributed} jointly models graph structure and node attributes in a two step graph generation process. AGM first exploits a structural generative graph model to compute underlying edge probabilities based on the structural properties of a given graph. It then learns attribute correlations among edges from the observed graph and combines them with the structural edge probabilities to sample edges conditioned on attribute values.
This process motivates us to uncover an underlying bias-free graph by separating out sensitive attributes and only conditioning on non-sensitive attributes for calculating edge probabilities.


\section{Preliminaries}
\label{sec:preliminary}
In this section, we first introduce our notations and general graph embedding concepts. 
Since the bias/fairness issues emerge most notably in prediction tasks involving humans, such as loan application or job recommendation, we will use user-related graphs as running examples to discuss our criterion for unbiased graph embedding. But we have to emphasize that this setting is only to illustrate the concept of unbiased graph embedding; and our proposed solution can be applied to any graph data and selected sensitive attributes to avoid biases in the learned embeddings.

\subsection{Notation}
Let $\gG = (\gV, \gE, \gA)$ be an undirected, attributed graph with a set of $N$ nodes $\gV$, a set of edges $\gE \subseteq \gV\times\gV$, and a set of $N$ attribute vectors $\gA$ (one attribute vector for each node). We use $(u,v)$ to denote an edge between node $u$ and node $v$. The number of attributes on each node is $K$, and $\gA = \{\va_1, \va_2, \dots, \va_N\}$, where $\va_u$ is a $K$-dimensional attribute value vector for node $u$. 
We assume all attributes are categorical and $\gS_i$ is the set of all possible values for attribute $i$. \footnote{We acknowledge that there are cases where attribute values are continuous, where discretization techniques can be applied.} For example, if node $u$ is a user node, and the $i$-th attribute is gender with possible values $\gS_i=\{\textit{Female},\textit{Male},\textit{Unknown}\}$, then $\va_u[i]=\textit{Female}$ indicates $u$ is a female. 
Without loss of generality, we assume the first $m$ attributes are sensitive, and $a_u[:m]$ and $a_u[m:]$ stands for the $m$ sensitive attributes and the rest of the attributes that are non-sensitive, respectively.


In the problem of graph embedding learning, we aim to learn an encoder $\mathrm{ENC}:\gV\rightarrow\sR^d$ that maps each node $u$ to a $d$-dimensional embedding vector $\vz_u=\mathrm{ENC}(u)$. We focus on the \textit{unsupervised} embedding setting which does not require node labels and the embeddings are learned via the \textit{link prediction task}. In this task, a scoring function $\rs_{\btheta}(\vz_u, \vz_v)$ with parameters $\btheta$ is defined to predict the probability of an edge $(u,v)\in\gE$ between node $u$ and node $v$ in the given graph. The loss for learning node embeddings and parameters of the encoder and scoring function is defined by:
\begin{equation}
\label{eq:loss}
    \sum_{(u,v)\in\gE} \gL_{edge}(\rs_{\btheta}(\vz_u, \vz_v)),
\end{equation}
where $\gL_{edge}$ is a per-edge loss function on $(u,v)\in\gE$. Such loss functions generally aim to maximize the likelihood of observed edges in the given graph, comparing to the negative samples of node pairs where edges are not observed \cite{mikolov2013distributed,grover2016node2vec}.

\subsection{Unbiased Graph Embedding}
\label{sec:unbiased}
Given a node $u$, we consider its embedding $\vz_u$ as unbiased with respect to an attribute $i$ if it is independent from the attribute.
Prior works evaluate such unbiasedness in the learned node embeddings by their ability to predict the values of the sensitive attributes \cite{bose2019compositional,palowitch2020monet,buyl2020debayes}. For example, they first train a classifier on a subset of node embeddings using their associated sensitive attribute values as labels. If the classifier cannot correctly predict the sensitive attribute values on the rest of node embeddings, one claims that the embeddings have low bias. If the prediction performance equals to that from random node embeddings, the learned embeddings are considered bias-free. In fact, such classifiers are often used as discriminators in adversarial methods where the classifier and the embeddings are learned jointly: the embeddings are pushed in directions where the classifier has low prediction accuracy \cite{madras2018learning,bose2019compositional}. 

There are also studies that use fairness measures such as demographic parity or equalized opportunity to define the unbiasedness of learned embeddings \cite{hardt2016equality,buyl2020debayes}. But we need to clarify that such fairness measures can only evaluate the fairness of the final prediction results for the intended downstream tasks, but cannot assess whether the embeddings are biased by, or contain any information about, sensitive attributes. In particular, fairness in a downstream task is only a necessary condition for unbiased embedding learning, not sufficient. The logic is obvious: unbiased embeddings can lead to fair prediction results as no sensitive attribute information is involved; but obtaining fairness in one task does not suggest the embeddings themselves are unbiased, e.g., those embeddings can still lead to unfair results in other tasks or even the fair results are obtained by other means, such as post-processing of the prediction results \cite{woodworth2017learning}. 
In \Cref{sec:exp}, we will use both the prediction accuracy on sensitive attributes and fairness measures on final tasks to evaluate the effectiveness of our unbiased graph embedding methods. 

\section{Effect of attributes in graph generation}
\label{sec:under-graph}
In this section, we discuss the generation of an observed graph by explicitly modeling the effects of node attributes in the process. In particular, we assume that there is an \emph{underlying structural graph} behind an observed graph, whose edge distribution is governed by the global graph structural properties such as degree distributions, diameter, and clustering coefficients.
The attributes in $\gA$ will modify the structural edge distribution based on effects like \emph{homophily} in social networks, where links are rewired based on the attribute similarities of the individuals \cite{mcpherson2001birds,fond2010randomization}. The modified edge distribution is then used to generate the observed graph. 

Formally, let $\gM$ be a structural generative graph model and $\Theta_{M}$ be the set of parameters that describe properties of the underlying structural graph. In particular, this set of parameters $\Theta_M$ is independent from node attributes in $\gA$. We consider the class of models that represent the set of possible edges in the graph as binary random variables $E_{uv},u\in\gV,v\in\gV$: i.e., the event $E_{uv}=1$ indicates $(u,v)\in\gE$. The model $\gM$ assigns a probability to $E_{uv}$ based on $\Theta_{M}$, $P_M(E_{uv}=1|\Theta_M)$. Therefore, the edges of an underlying structural graph $\gG_M$ can be considered as samples from $Bernoulli(P_M(E_{uv}=1|\Theta_M))$. There are many such structural models $\gM$ such as the Chung Lu model \cite{Chung2002average} and Kronecker Product Graph Model \cite{leskovec2010kronecker}. Note that $\gM$ does not consider node attributes in the generation of the structural graph.  

Now we involve the attributes in the generative process. Let $C\in\{(\va_i,\va_j)|\,i\in\gV,j\in\gV\}$ be a variable indicating the \emph{attribute value combination} of a randomly sampled pair of nodes, which is independent from $\Theta_M$. Note that $C$ instantiated by different node pairs can be the same, as different nodes can have the same attribute values. The conditional probability of an edge between $u$ and $v$, given the corresponding attribute values on them and the structural parameters $\Theta_M$, is $P_{o}(E_{uv}=1|C=\va_{uv},\Theta_M)$, where $\va_{uv}=(\va_u, \va_v)$ denotes the attribute value combination on nodes $u$ and $v$. Based on Bayes' Theorem, we have
\begin{align}
\label{eq:bayes}
    & P_o(E_{uv}=1|C=\va_{uv},\Theta_M)  \\\nonumber
  =\,& \frac{P_o(C=\va_{uv}|E_{uv}=1,\Theta_M)P_o(E_{uv}=1|\Theta_M)}{P_o(C=\va_{uv}|\Theta_M)} \\\nonumber
  =\,& P_M(E_{uv}=1|\Theta_M)\frac{P_o(C=\va_{uv}|E_{uv}=1,\Theta_M)}{P_o(C=\va_{uv}|\Theta_M)},\forall u\in\gV, \forall v\in\gV,
\end{align}
where the prior distribution on $E_{uv}$ is specified by the structural model $\gM$: i.e., $P_o(E_{uv}=1|\Theta_M) = P_M(E_{uv}=1|\Theta_M)$, and the posterior distribution accounts for the influences from the attribute value combinations. Therefore, the edge probabilities used to generate the observed graph with node attributes is a \emph{modification} of those from a structural graph defined by $\gM$ and $\Theta_M$. 
It is important to clarify that the node attributes are given ahead of graph generation. They are the input to the generative process, not the output. Hence, $P_o(C=\va_{uv}|E_{uv}=1,\Theta_M)$ represents the probability that in all edges, the specific attribute value combination $\va_{uv}$ is observed on an edge's incident nodes. It 
is thus the same for all edges whose incident nodes have the same attribute value combination.


To simplify the notation, let us define a function that maps the attribute value combination $\va_{uv}$ to the probability ratio that modifies the structural graph into the observed graph by
\begin{equation*}
    R(\va_{uv}) \coloneqq \frac{P_o(C=\va_{uv}|E_{uv}=1,\Theta_M)}{P_o(C=\va_{uv}|\Theta_M)}, \forall u\in\gV, \forall v\in\gV.
\end{equation*}
Thus we can rewrite \eqref{eq:bayes} by 
\begin{equation}
\label{eq:prob-decomp}
    P_o(E_{uv}=1|C=\va_{uv},\Theta_M) = P_M(E_{uv}=1|\Theta_M)R(\va_{uv}). 
\end{equation}
In this way, we explicitly model the effect of node attributes by $R(\va_{uv})$, which modifies the structural graph distribution $P_M(E_{uv}=1|\Theta_M)$ for generating the observed graph $\gG$. 

\begin{figure}
    \centering
    \includegraphics[width=0.85\linewidth]{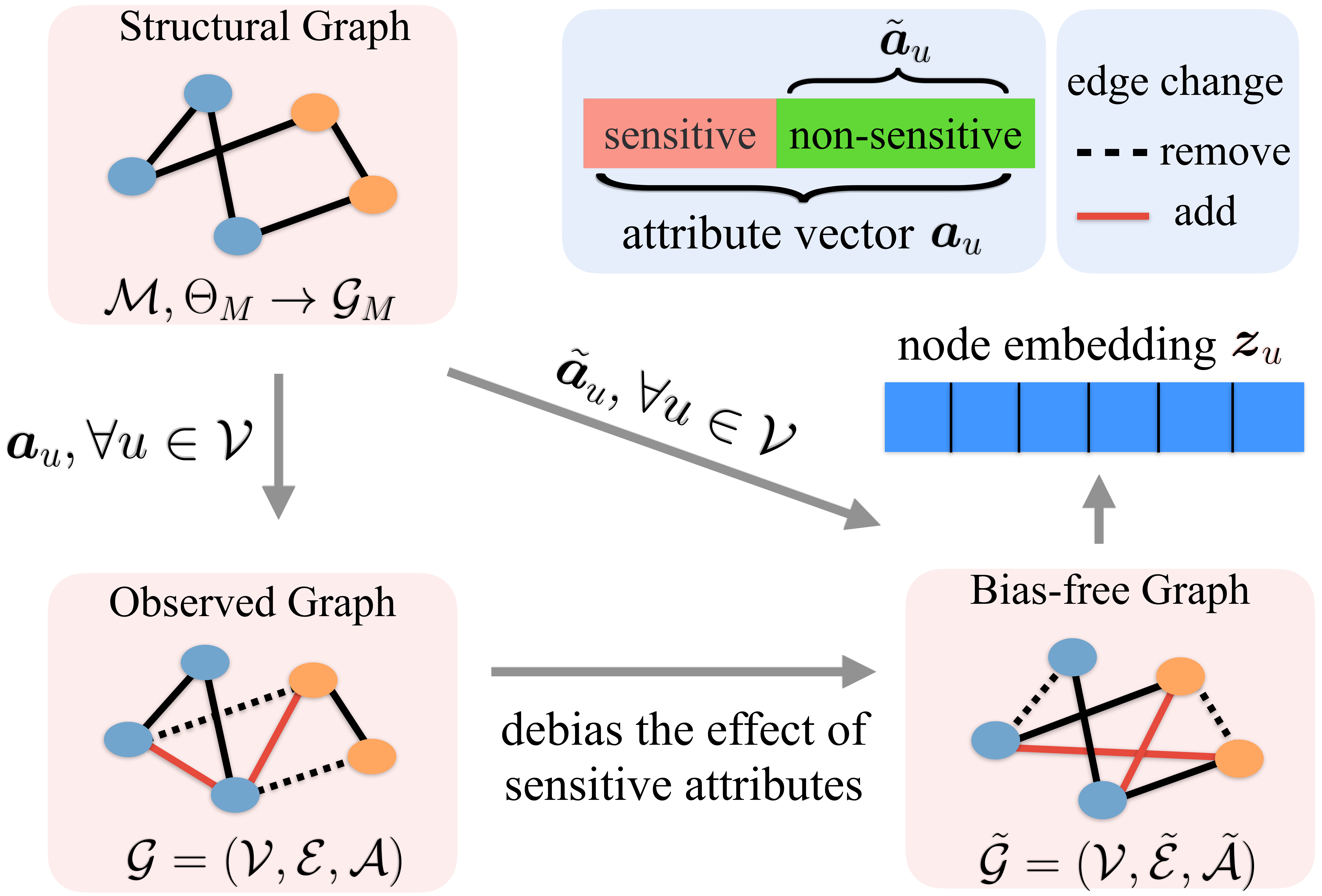}
    \vspace{-3mm}
    \caption{Illustration of Unbiased Graph Embedding (\UGE). The color of the nodes represents the value of their attributes, and different line styles suggest how the observed edges are influenced by attributes in the generative process.}
    \label{fig:UGE}
    \vspace{-2mm}
\end{figure}
\section{Unbiased Graph Embedding from a Bias-Free Graph}
\label{sec:UGE}
In this section, we describe our proposed methods for learning unbiased node embeddings based on the generative modeling of the effects of sensitive attributes in \Cref{sec:under-graph}. In a nutshell, we aim to get rid of the sensitive attributes and modify the structural edge probabilities by only conditioning on non-sensitive attributes. This gives us the edge probabilities of a bias-free graph, from which we can learn unbiased node embeddings.
We illustrate this principle in \Cref{fig:UGE}. 
Consider a world without the sensitive attributes, and the attribute vector of node $u$ becomes $\tilde\va_u = \va_u[m:]$, which only include non-sensitive attributes in $\va_u$. We denote $\tilde{\gG}=(\gV, \tilde\gE, \tilde\gA)$ as the corresponding new graph generated with $\tilde\va_u, \forall u\in\gV$, and $\tilde\va_{uv}=(\tilde\va_u, \tilde\va_v)$. Therefore, $\tilde\gG$ is a bias-free graph without influence from sensitive attributes. If we can learn node embeddings from $\tilde\gG$ instead of $\gG$, the embeddings are guaranteed to be unbiased with respect to sensitive attributes. Specifically, the edge probabilities 
used for generating $\tilde\gG$ can be written as
\begin{equation}
\label{eq:prob-decomp2}
    P_{\tilde o}(E_{uv}=1|\tilde{C}=\tilde\va_{uv},\Theta_M) = P_M(E_{uv}=1|\Theta_M)\tilde R(\tilde\va_{uv}), 
\end{equation}
where
\begin{equation}
    \tilde R(\tilde\va_{uv}) \coloneqq \frac{P_{\tilde o}(\tilde{C}=\tilde\va_{uv}|E_{uv}=1,\Theta_M)}{P_{\tilde o}(\tilde{C}=\tilde\va_{uv}|\Theta_M)}, \forall u\in\gV, \forall v\in\gV,
\end{equation}
$\tilde{C}\in\{(\tilde\va_i,\tilde\va_j)|i\in\gV,j\in\gV\}$ is the random variable for attribute value combinations without sensitive attributes, and $P_{\tilde o}$ indicates the distributions used in generating $\tilde\gG$. We name the class of methods that learn embeddings from $\tilde\gG$ as \UGE, simply for \textbf{U}nbiased \textbf{G}raph \textbf{E}mbedding. 
Next we introduce two instances of \UGE. The first is \UGE-W, which reweighs the per-edge loss such that the total loss is from $\tilde\gG$ in expectation. The second method is \UGE-R, which adds a regularization term to shape the embeddings to satisfy the properties as those directly learned from $\tilde\gG$.

\subsection{Weighting-Based \UGE}
To compose a loss based on $\tilde\gG$, we modify the loss function in \eqref{eq:loss} by reweighing the loss term on each edge as
\begin{equation}
\label{eq:uge-loss}
    \gL_{\UGE-W}(\gG) = \sum_{(u,v)\in\gE} \gL_{edge}(\rs_{\btheta}(\vz_u, \vz_v))\frac{\tilde R(\tilde\va_{uv})}{R(\va_{uv})}.
\end{equation}
The following theorem shows that, in expectation, this new loss is equivalent to the loss for learning node embeddings from $\tilde\gG$.

\begin{theorem}
Given a graph $\gG$, and $\tilde R(\tilde\va_{uv}) / R(\va_{uv}), \forall (u,v)\in\gE$,  $\gL_{\UGE-W}(\gG)$ is an unbiased loss with respect to $\tilde\gG$.
\end{theorem}
\begin{proof} We take expectation over the edge observations in $\gG$ as
\begin{align}
\label{eq:expectation}
    & \E\big[\gL_{\UGE-W}(\gG)\big] \\\nonumber
    =\,& \E\Bigg[\sum_{(u,v)\in\gE} \gL_{edge}(\rs(\vz_u, \vz_v))\frac{\tilde R(\tilde\va_{uv})}{R(\va_{uv})}\Bigg] \\\nonumber
    =\,& \E\Bigg[\sum_{u\in\gV, v\in\gV} \gL_{edge}(\rs(\vz_u, \vz_v))\frac{\tilde R(\tilde\va_{uv})}{R(\va_{uv})}\cdot E_{uv}\Bigg] \\\nonumber
    =\,& \sum_{u\in\gV, v\in\gV} \gL_{edge}(\rs(\vz_u, \vz_v))\frac{\tilde R(\tilde\va_{uv})}{R(\va_{uv})} \cdot P_o(E_{uv}=1|C=\va_{uv},\Theta_M) \\\nonumber  
    *=\,& \sum_{u\in\gV, v\in\gV} \gL_{edge}(\rs(\vz_u, \vz_v))\cdot P_{\tilde o}(E_{uv}=1|\tilde{C}=\tilde\va_{uv},\Theta_M) \\\nonumber
    =\,& \E\Bigg[\sum_{(u,v)\in\tilde\gE} \gL_{edge}(\rs(\vz_u, \vz_v))\Bigg].
\end{align}
The step marked by $*$ uses \eqref{eq:prob-decomp} and \eqref{eq:prob-decomp2}.
\end{proof}
\UGE-W is closely related to the idea of importance sampling \cite{kloek1978bayesian}, which analyzes the edge distribution of the bias-free graph $\tilde\gG$ by observations from the given graph $\gG$. The only thing needed for deploying \UGE-W in existing graph embedding methods is to calculate the weights $\tilde R(\tilde\va_{uv})/R(\va_{uv})$. 
To estimate $R(\va_{uv})$, we need the estimates of $P_o(C=\va_{uv}|E_{uv}=1,\Theta_M)$ and $P_o(C=\va_{uv}|\Theta_M)$. With maximum likelihood estimates on the observed graph, we have 
\begin{align}
\label{eq:estimate-R}
    P_o(C=\va_{uv}|E_{uv}=1,\Theta_M) &\approx \frac{\sum_{(i,j)\in\gE}\sI[\va_{ij}=\va_{uv}]}{|\gE|}, \\
    P_o(C=\va_{uv}|\Theta_M) &\approx \frac{\sum_{i\in\gV,j\in\gV}\sI[\va_{ij}=\va_{uv}]}{N^2}.
\end{align}
Similarly we can estimate $\tilde R(\tilde \va_{uv})$ by
\begin{align}
    P_{\tilde o}(\tilde{C}=\tilde\va_{uv}|E_{uv}=1,\Theta_M) &\approx \frac{\sum_{(i,j)\in\tilde\gE}\sI[\tilde\va_{ij}=\tilde\va_{uv}]}{|\tilde\gE|}, \\\label{eq:estimate-tildeR}
    P_{\tilde o}(\tilde{C}=\tilde\va_{uv}|\Theta_M) &\approx \frac{\sum_{i\in\gV,j\in\gV}\sI[\tilde\va_{ij}=\tilde\va_{uv}]}{N^2}.
\end{align}
Note that the estimation of $P_{\tilde o}(\tilde{C}=\tilde\va_{uv}|E_{uv}=1,\Theta_M)$ is based on $\tilde\gE$, which is unfortunately from the implicit bias-free graph $\tilde\gG$ and unobservable. But we can approximate it with $\gE$ in the following way: after grouping node pairs by non-sensitive attribute value combinations $\tilde\va_{uv}$, the sensitive attributes only re-route the edges but do not change the number of edges in each group. Thus,
\begin{align}
    P_{\tilde o}(\tilde{C}=\tilde\va_{uv}|E_{uv}=1,\Theta_M) &\approx \frac{\sum_{(i,j)\in\tilde\gE}\sI[\tilde\va_{ij}=\tilde\va_{uv}]}{|\tilde\gE|} \\\nonumber
    &= \frac{\sum_{i\in\gV,j\in\gV,\tilde\va_{ij}=\tilde\va_{uv}}\sI[(i,j)\in\tilde\gE]}{|\tilde\gE|} \\\nonumber
    &= \frac{\sum_{i\in\gV,j\in\gV,\tilde\va_{ij}=\tilde\va_{uv}}\sI[(i,j)\in\gE]}{|\tilde\gE|} \\\nonumber
    &= \frac{\sum_{(i,j)\in\gE}\sI[\tilde\va_{ij}=\tilde\va_{uv}]}{|\gE|}.
\end{align}

For node pairs with the same attribute value combination, \eqref{eq:estimate-R}-\eqref{eq:estimate-tildeR} only need to be calculated once instead of for each pair. This can be done by first grouping node pairs by their attribute value combinations and then perform estimation in each group. However, when there are many attributes or attributes can take many unique values, the estimates may become inaccurate since there will be many groups and each group may only have a few nodes. In this case, we can make independence assumptions among the attributes. 
For example, by assuming they are independent, the estimate for a specific attribute value combination over all the $K$ attributes becomes the product of $K$ estimates, one for each attribute. The non-sensitive attributes can be safely removed under this assumption with $\tilde R(\tilde\va_{uv})=1$, and only $R(\va_{uv})$ needs to be estimated as $R(\va_{uv})=\prod_{i=1}^m R(\va_{uv}[i])$.
Since \UGE-W only assigns pre-computed weights to the loss, the optimization based on it will not increase the complexity of any graph embedding method. 

\subsection{Regularization-Based \UGE}
We propose an alternative way for \UGE which adds a regularization term to the loss function that pushes the embeddings to satisfy properties required by the bias-free graph $\tilde\gG$. Specifically, when the node embeddings are learned from $\tilde\gG$, their produced edge distributions should be the same with and without the sensitive attributes. To enforce this condition, we need to regularize the discrepancy between $P_o(E_{uv}=1|C=\va_{uv},\Theta_M)$ and $P_{\tilde o}(E_{uv}=1|\tilde{C}=\tilde\va_{uv},\Theta_M)$ induced from the node embeddings. We can use the scores in $\rs_{\btheta}(\vz_u, \vz_v)$ as a proxy to represent edge probability produced by the embeddings of nodes $u$ and $v$, i.e., high $\rs_{\btheta}(\vz_u, \vz_v)$ indicates high probability of an edge between $u$ and $v$. 
We can measure $P_o(E_{uv}=1|C=\va_{uv},\Theta_M)$ by aggregating node pairs with the same attribute value combination to marginalize out the effect of $\Theta_M$ and focus on the influence from attributes as
\begin{equation}
     Q_{\va_{uv}} = \frac{1}{N_{\va_{uv}}}\sum_{i\in\gV,j\in\gV,\va_{ij}=\va_{uv}}\rs_{\btheta}(\vz_i, \vz_j),
\end{equation}
where we use $Q_{\va_{uv}}$ to denote the approximated measure of $P_o(E_{uv}=1|C=\va_{uv},\Theta_M)$, and $N_{\va_{uv}}$ is the number of node pairs that has the attribute value combination $\va_{uv}$. For pairs with the same attribute value combination, $Q_{\va_{uv}}$ only needs to be calculated once. Similarly, $P_{\tilde o}(E_{uv}=1|\tilde{C}=\tilde\va_{uv},\Theta_M)$ can be represented by $Q_{\tilde\va_{uv}}$, which can be obtained by aggregating the scores over pairs with non-sensitive attribute value combination $\tilde\va_{uv}$.
Finally, we use $\ell_2$ distance between $Q_{\va_{uv}}$ and $Q_{\tilde\va_{uv}}$ as the regularization
\begin{align}
\label{eq:uge-r}
    &\gL_{\UGE-R}(\gG) \\\nonumber
    =&\; \sum_{(u,v)\in\gE} \gL_{edge}(\rs_{\btheta}(\vz_u, \vz_v)) + \lambda\sum_{u\in\gV, v\in\gV}\norm{Q_{\va_{uv}} - Q_{\tilde\va_{uv}}}_2,
\end{align}
where $\lambda$ controls the trade-off between the per-edge losses and the regularization.

In contrast to adversarial regularizations employed in prior work \cite{madras2018learning,bose2019compositional,agarwal2021towards, dai2021say}, \UGE-R takes a different perspective in regularizing the discrepancy between graphs with and without sensitive attributes induced from the embeddings. All previous regularization-based methods impose the constraint on individual edges. 
We should note that the regularization term is summed over all node pairs, which has a complexity of $O(N^3)$ and can be costly to calculate. But in practice, we can add the regulariztaion by only sampling  batches of node pairs in each iteration during model update, and use $\lambda$ to compensate the strength of the regularization.

\subsection{Combined Method}
\label{sec:discuss-uge}

As hinted in \cref{sec:intro}, \UGE-W is a sufficient condition for unbiased graph embedding, since it directly learns node embeddings from a bias-free graph. \UGE-R is a necessary condition, as it requires the learned embeddings to satisfy the properties of a bias-free graph. We can combine them to trade-off the debiasing effect and utility,
\begin{align}
    &\gL_{\UGE-C}(\gG) \\\nonumber
    =&\; \sum_{(u,v)\in\gE} \gL_{edge}(\rs_{\btheta}(\vz_u, \vz_v))\frac{\tilde R(\tilde\va_{uv})}{R(\va_{uv})} + \lambda\sum_{u\in\gV, v\in\gV}\norm{Q_{\va_{uv}} - Q_{\tilde\va_{uv}}}_2,
\end{align}
where we use $\gL_{\UGE-C}(\gG)$ to represent the combined method.
$\gL_{\UGE-C}(\gG)$ thus can leverage the advantages of both \UGE-W and \UGE-R to achieve better trade-offs between the unbiasedness and the utility of node embeddings in downstream tasks. 

\section{Experiments}
\label{sec:exp}
In this section, we study the empirical performance of \UGE on three benchmark datasets in comparison to several baselines. In particular, we apply \UGE to \emph{five} popularly adopted backbone graph embedding models to show its wide applicability. To evaluate the debiasing performance, the node embeddings are firstly evaluated by their ability to predict the value of sensitive attributes, where lower prediction performance means better debiasing effect. Then a task-specific metric is used to evaluate the utility of the embeddings. Besides, we also apply fairness metrics in the link prediction results to demonstrate the potential of using embeddings from \UGE to achieve fairness in downstream tasks. 

\begin{table}
 \caption{Statistics of evaluation graph datasets.}
 \vspace{-2mm}
  \label{tab:data}
  \begin{tabular}{lrrr}
    \toprule
    \textbf{Statistics} & \textbf{Pokec-z} & \textbf{Pokec-n} & \textbf{MovieLens-1M} \\
    \midrule
    \# of nodes & $67,796$ & $66,569$ & $9,992$ \\
    \# of edges & $882,765$ & $729,129$ & $1,000,209$ \\
    Density & $0.00019$ & $0.00016$ & $0.01002$ \\
    \bottomrule
  \end{tabular}
  \vspace{-4mm}
\end{table}

\begin{table*}[]
\centering
\caption{Unbiasedness evaluated by Micro-F1 on Pokec-z and Pokec-n. Bold numbers highlight the best in each row.}
\label{tab:pokec-results}
\begin{tabular}{cccccccccc}
\toprule
Dataset & Embedding Model & Prediction Target & No Debiasing & Fairwalk & CFC & UGE-W & UGE-R & UGE-C & Random \\ 
\midrule
\multirow{3}{*}{Pokec-z} & \multirow{3}{*}{GAT} & Gender (Micro-F1) & 0.6232 & 0.6135 & 0.5840 & 0.6150 & 0.6094 & \textbf{0.5747} & 0.4921 \\
 & & Region (Micro-F1) & 0.8197 & 0.8080 & 0.7217 & 0.6784 & 0.7660 & \textbf{0.6356} & 0.4966 \\
 & & Age (Micro-F1) & 0.0526 & 0.0522 & 0.0498 & 0.0431 & 0.0545 & \textbf{0.0429} & 0.0007 \\
\midrule
\multirow{3}{*}{Pokec-n} & \multirow{3}{*}{node2vec} & Gender (Micro-F1) & 0.5241 & 0.5291 & 0.5241 & 0.5187 & \textbf{0.5095} & 0.5158 & 0.5078 \\
 & & Region (Micro-F1) & 0.8690 & 0.8526 & 0.8423 & 0.8158 & 0.6975 & \textbf{0.6347} & 0.4987 \\
 & & Age (Micro-F1) & 0.0626 & 0.0534 & 0.0426 & 0.0305 & 0.0294 & \textbf{0.0194} & 0.0002 \\ \bottomrule
\end{tabular}
\end{table*}

\subsection{Setup}
\noindent\textbf{$\bullet$ Dataset.} 
We use three public user-related graph datasets, Pokec-z, Pokec-n and MovieLens-1M, where the users are associated with sensitive attributes to be debiased. The statistics of these three datasets are summarized in Table \ref{tab:data}.
Pokec\footnote{https://snap.stanford.edu/data/soc-pokec.html} is an online social network in Slovakia, which contains anonymized data of millions of users \cite{takac2012data}. Based on the provinces where users belong to, we used two sampled datasets named as \textbf{Pokec-z} and \textbf{Pokec-n} adopted from \cite{dai2020learning}, which consist of users belonging to two major regions of the corresponding provinces, respectively. In both datasets, each user has a rich set of features, such as education, working field, interest, etc.; and we include \textit{gender, region} and \textit{age} as (sensitive) attributes whose effect will be studied in our evaluation.
\textbf{MovieLens-1M}\footnote{https://grouplens.org/datasets/movielens/1m/} is a popular movie recommendation benchmark, which contains around one million user ratings on movies \cite{harper2015movielens}. In our experiment, we construct a bipartite graph which consists of user and movie nodes and rating relations as edges. The dataset includes \textit{gender, occupation} and \textit{age} information about users, which we treat as sensitive attributes to be studied. We do not consider movie attributes, and thus when applying \UGE, only user attributes are counted for our debiasing purpose.

\noindent\textbf{$\bullet$ Graph embedding models.}
\UGE is a general recipe for learning unbiased node embeddings, and can be applied to different graph embedding models. We evaluate its effectiveness on five representative embedding models in the supervised setting with the link prediction task. \textbf{GCN} \cite{kipf2016semi}, \textbf{GAT} \cite{velivckovic2017graph}, \textbf{SGC} \cite{wu2019simplifying} and \textbf{node2vec} \cite{grover2016node2vec} are deep learning models, and we use dot product between two node embeddings to predict edge probability between them and apply cross-entropy loss for training. 
\textbf{MF} \cite{mnih2008probabilistic} applies matrix factorization to the adjacency matrix. Each node is represented by an embedding vector learned with pairwise logistic loss \cite{rendle2012bpr}.

\noindent\textbf{$\bullet$ Baselines.}
We consider three baselines for generating unbiased node embeddings.  
(1) \textbf{Fairwalk} \cite{rahman2019fairwalk} is based on node2vec, which modifies the pre-processing of random-walk generation by grouping neighboring nodes with their values of the sensitive attributes. Instead of randomly jumping to a neighbor node, Fairwalk firstly jumps to a group and then sample a node from that group for generating random walks. We extend it to GCN, GAT and SGC by sampling random walks of size $1$ to construct the corresponding per-edge losses for these embedding models. 
(2) \textbf{Compositional Fairness Constraints (CFC)} \cite{bose2019compositional} is an algorithmic method, which adds an adversarial regularizer to the loss by jointly training a composition of sensitive attribute discriminators. We apply CFC to all graph embedding models and tune the weight on the regularizer, where larger weights are expected to result in embeddings with less bias but lower utility.
(3) \textbf{Random} embeddings are considered as a bias-free baseline. We generate random embeddings by uniformly sampling the value of each embedding dimension from [0, 1].

It is worth mentioning that a recent work \textbf{DeBayes} \cite{buyl2020debayes}, which is based on the conditional network embedding (CNE) \cite{kang2018conditional}, includes the sensitive information in a biased prior for learning unbiased node embeddings. We did not include it since it is limited to CNE and cannot be easily generalized to other graph embedding models. Besides, we found the bias prior calculation in DeBayes does not scale to large graphs where the utility of resulting node embeddings is close to random. The original paper \cite{buyl2020debayes} only experimented with two small graph datasets with less than $4K$ nodes and $100K$ edges. By default, \UGE follows Fairwalk to debias each of the sensitive attributes separately in experiments without independence assumption between attributes. CFC debiases all sensitive attributes jointly as suggested in the original paper.\footnote{\UGE can debias either a single attribute or multiple attributes jointly by removing one or more attributes in the bias-free graph.}

\noindent\textbf{$\bullet$ Configurations.}
For the Pokec-z and Pokec-n datasets, we apply GCN, GAT, SGC and node2vec as embedding models and apply debiasing methods on top of them. For each dataset, we construct positive examples for each node by collecting $N_{pos}$ neighboring nodes with $N_{pos}$ equal to its node degree, and randomly sample $N_{neg}=20\times N_{pos}$ unconnected nodes as negative examples.
For each node, we use $90\%$ positive and negative examples for training and reserve the rest $10\%$ for testing.
For Movielens-1M, we follow common practices and use MF as the embedding model \cite{rahman2019fairwalk,bose2019compositional}. We do not evalaute Fairwalk on this dataset since there is no user-user connections and fair random walk cannot be directly applied. The rating matrix is binarized to create a bipartite user-movie graph for MF. We use $80\%$ ratings for training and $20\%$ for testing. For all datasets and embedding models, we set the node embedding size to $d=16$. We include more details about model implementations and hyper-parameter tuning in \Cref{appendix:settings}.

In \Cref{sec:unbiasedness}, we compare the unbiasedness and utility of embeddings from different baselines. We evaluate fairness resulted from the embeddings in \Cref{sec:fairness}. We study the unbiasedness-utility trade-off in \UGE and CFC in \Cref{sec:reg-weights}. Since there is a large number of experimental settings composed of different datasets, embedding models, and baselines, we report results from different combinations in each section to maximize the coverage in each component, and include the other results in \Cref{appendix:results}.

\begin{figure}[t]
 \centering
 \begin{subfigure}[t]{\linewidth}
    \centering
    \includegraphics[width=\linewidth]{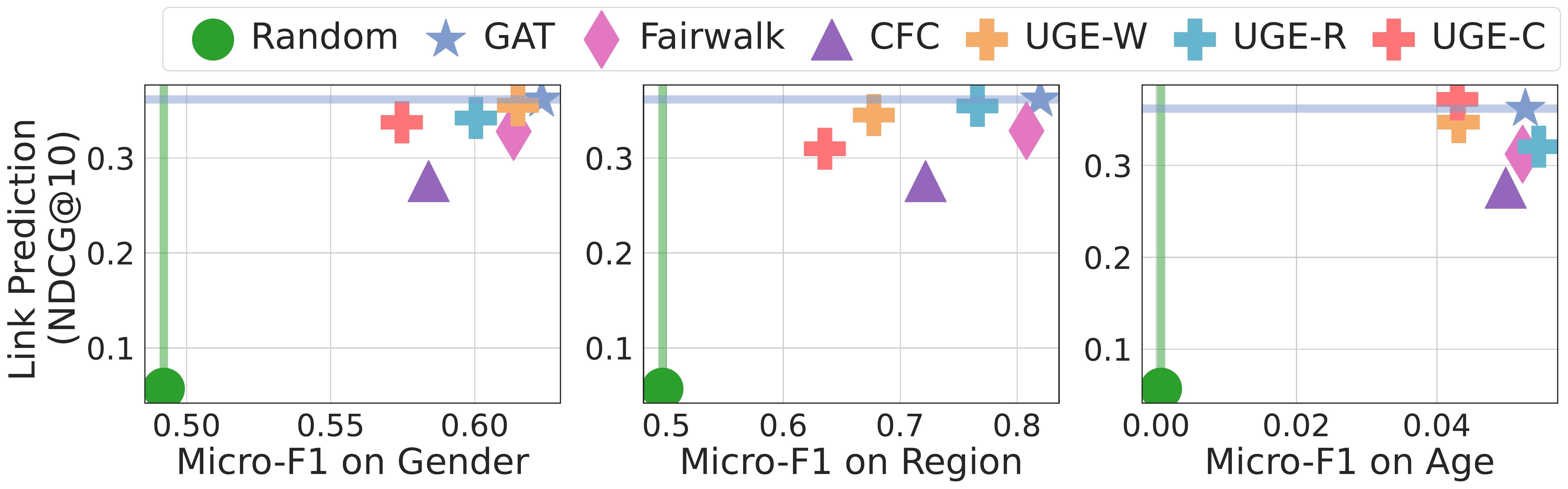}
    \caption{Pokec-z with GAT as embedding model}
    \label{fig:tradeoff-pokec-z}
 \end{subfigure}
 \begin{subfigure}[t]{\linewidth}
    \centering
    \includegraphics[width=\linewidth]{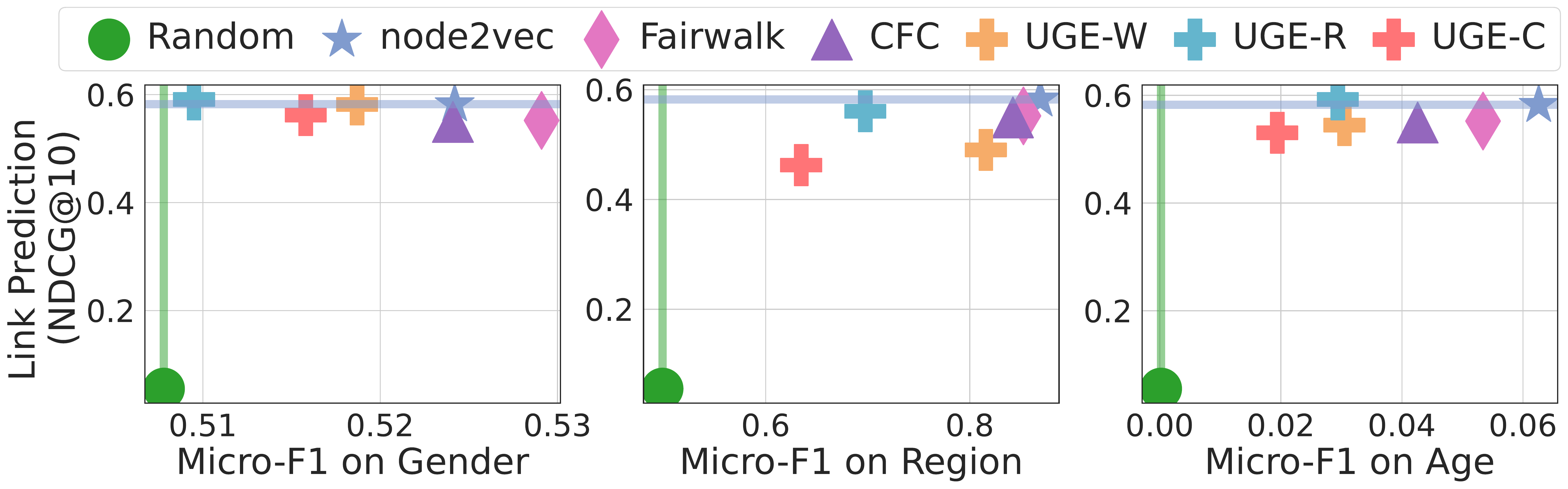}
    \caption{Pokec-n with node2vec as embedding model} 
    \label{fig:tradeoff-pokec-n}
 \end{subfigure}
 \begin{subfigure}[t]{\linewidth}
    \centering
    \includegraphics[width=\linewidth]{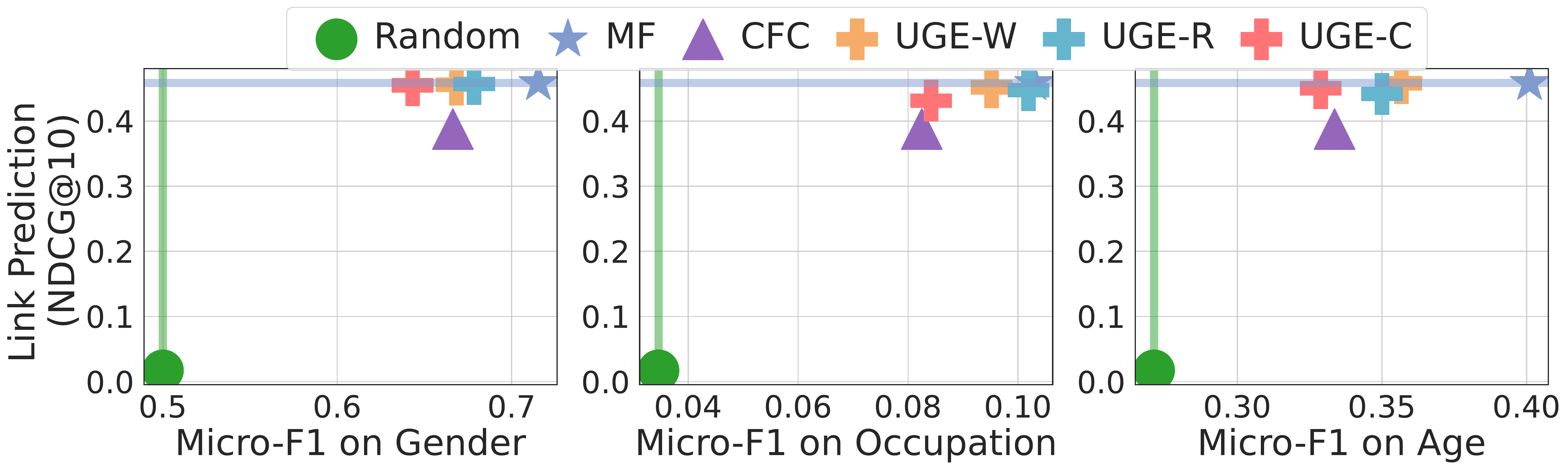}
    \caption{MovieLens-1M with MF as embedding model} 
    \label{fig:tradeoff-movielens}
 \end{subfigure}
 \setlength{\abovecaptionskip}{-1pt}
 \caption{Trade-off between the utility (by NDCG@10) and unbiasedness (by Micro-F1) of different methods. Random embeddings give the lowest Micro-F1 (green line), and no debiasing gives the best NDCG@10 (blue line). 
 An ideal debiasing method should locate itself at the upper left corner.
 }
 \label{fig:tradeoff}
\end{figure}

\subsection{Unbiasedness and Utility Trade-off}
\label{sec:unbiasedness}
We firstly compare the unbiasedness of node embeddings from different debiasing methods. For each sensitive attribute, we train a logistic classifier with 80\% of the nodes using their embeddings as features and attribute values as labels. We then use the classifier to predict the attribute values on the rest of 20\% nodes and evaluate the performance with Micro-F1. 
The Micro-F1 score can be used to measure the severity of bias in the embeddings, i.e., a lower score means lower bias in the embeddings. Random embeddings are expected to have the lowest Micro-F1 and embeddings without debiasing should have the highest Micro-F1. We show the results on Pokec-z with GAT as base embedding model and Pokec-n with node2vec as the base embedding model in \Cref{tab:pokec-results}.
From the results, we see that embeddings from \UGE methods always have the least bias against all baselines with respect to all sensitive attributes and datasets. This confirms the validity of learning unbiased embeddings from a bias-free graph. Besides, by combining \UGE-W and \UGE-R, \UGE-C usually produces the best debiasing effect, which demonstrates the complementary effect of the two methods. 

Besides the unbiasedness, the learned embeddings need to be effective when applied to downstream tasks. In particular, we use NDCG@10 evaluated on the link prediction task to measure the utility of the embeddings. Specifically, for each target node, we create a candidate list of 100 nodes that includes all its observed neighbor nodes in the test set and randomly sampled negative nodes. Then NDCG@10 is evaluated on this list with predicted edge probabilities from the node embeddings. \Cref{fig:tradeoff-pokec-z,fig:tradeoff-pokec-n} show the unbiasedness as well as the utility of embeddings from different methods in correspondence to the two datasets and embedding models in \Cref{tab:pokec-results}. \Cref{fig:tradeoff-movielens} shows the results on MovieLens-1M with MF as the embedding model.

In these plots, 
different embedding methods are represented by different shapes in the figures, and we use different colors to differentiate \UGE-W, \UGE-R and \UGE-C. Random embeddings do not have any bias and provide the lowest Micro-F1 (green line), while embeddings without any debiasing gives the highest NDCG@10 (blue line). To achieve the best utility-unbiasedness trade-off, an ideal debiasing method should locate itself at the upper left corner. As shown in the figures, \UGE based methods achieve the most encouraging trade-offs on these two contradicting objectives in most cases.
\UGE-C can usually achieve better debiasing effect, without sacrificing too much utility. \UGE-W and \UGE-R maintain high utility but are less effective than the combined version. CFC can achieve descent unbiasedness in embeddings, but the utility is seriously compromised (such as in Pokec-z and MovieLens-1M). Fairwalk unfortunately does not present an obvious debiasing effect.

\begin{figure}[t]
 \centering
  \includegraphics[width=0.95\linewidth]{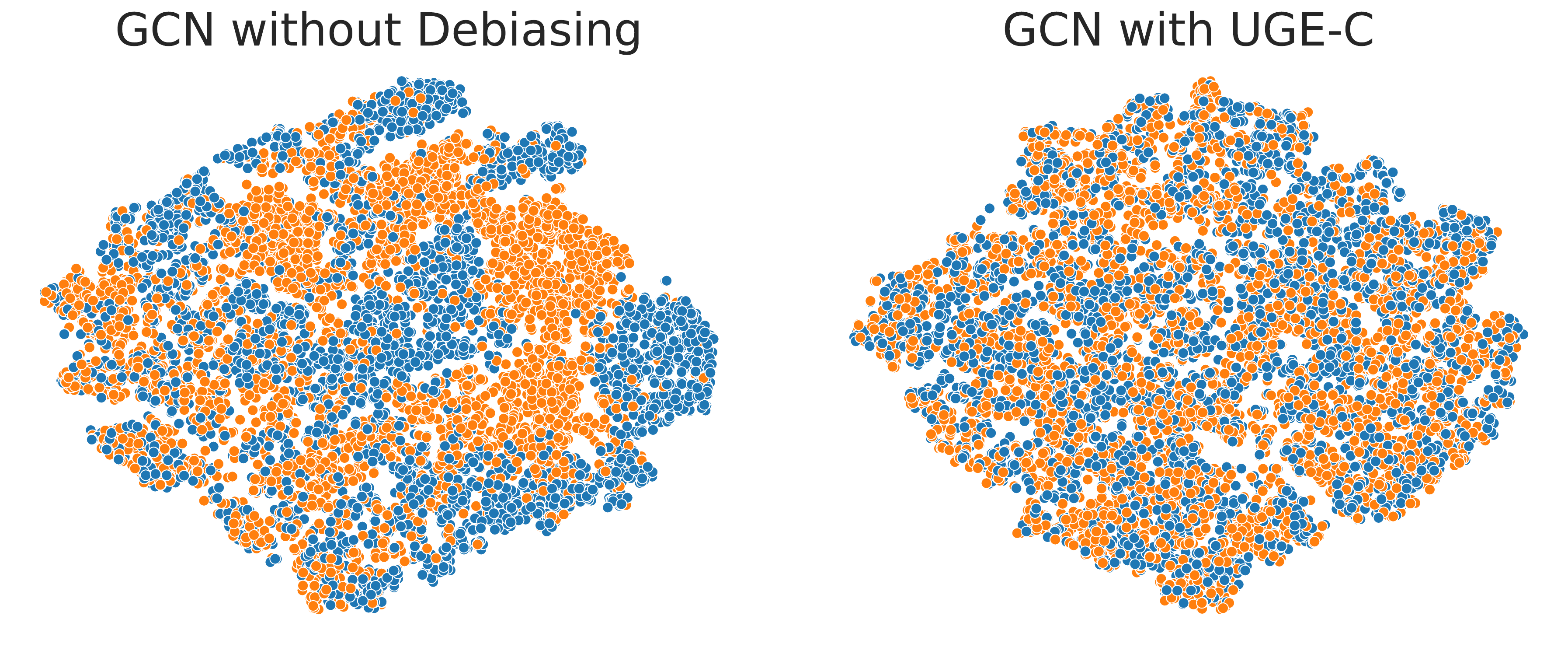}
  \setlength{\abovecaptionskip}{-1pt}
 \caption{Visualization of embeddings learned on Pokec-n. Node color represents the region of the nodes.}
 \label{fig:illustrate-pokec-n}
\end{figure}

To further visualize the debiasing effect of \UGE, we use t-SNE to project the node embeddings on Pokec-n to a 2-D space in \Cref{fig:illustrate-pokec-n}. 
The left plot shows the embeddings learned via GCN without debiasing, and the right plot exhibits the debiased embeddings by applying \UGE-C on GCN to debias the region attibute. Node colors represent the region value. Without debiasing, the embeddings are clearly clustered to reflect the regions of nodes. With \UGE-C, embeddings from different regions are blended together, showing the effect of removing the region information from the embeddings. 

\begin{figure}[t]
 \centering
 \begin{subfigure}[b]{0.88\linewidth}
 \centering
 \includegraphics[width=\linewidth]{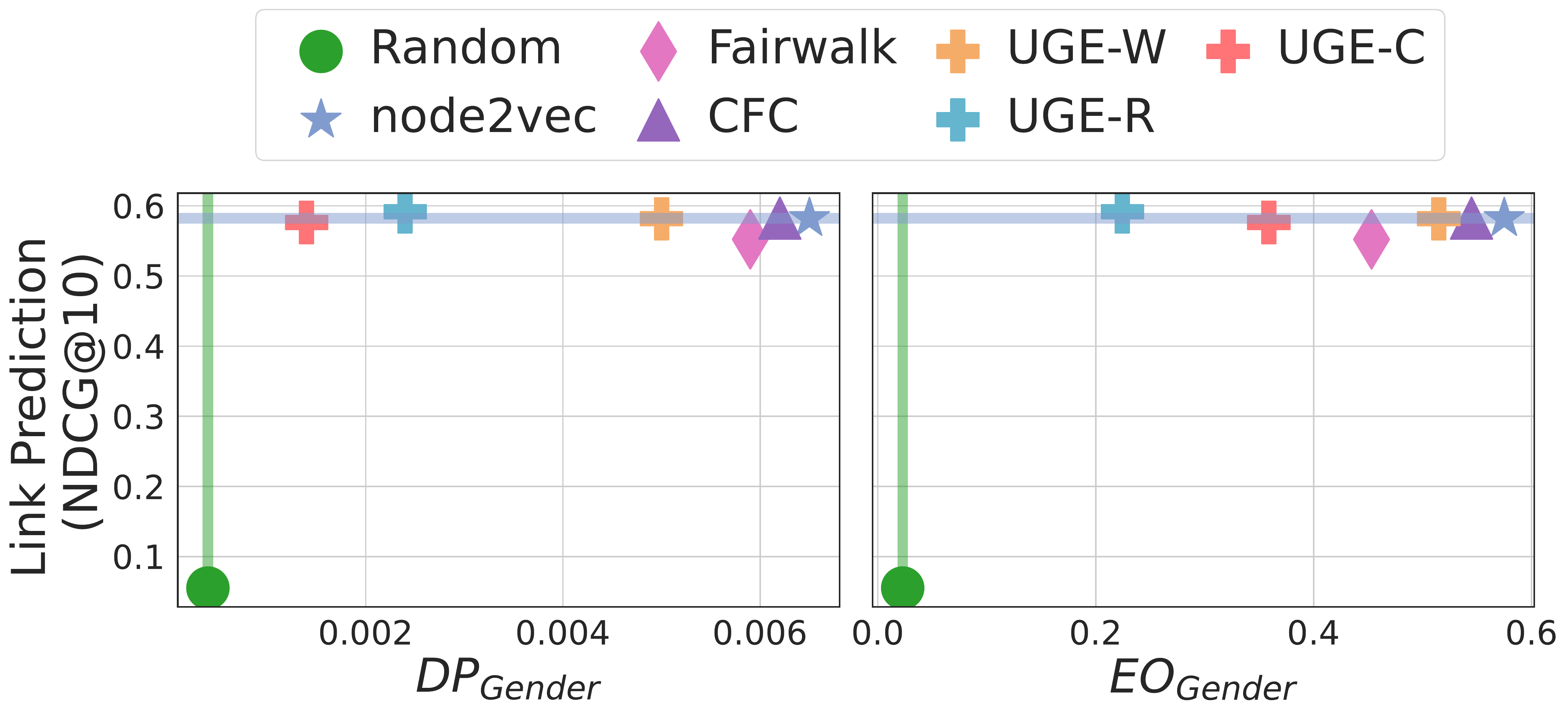}
 \end{subfigure}
 \hfill
 \begin{subfigure}[b]{0.88\linewidth}
 \centering
 \includegraphics[width=\linewidth]{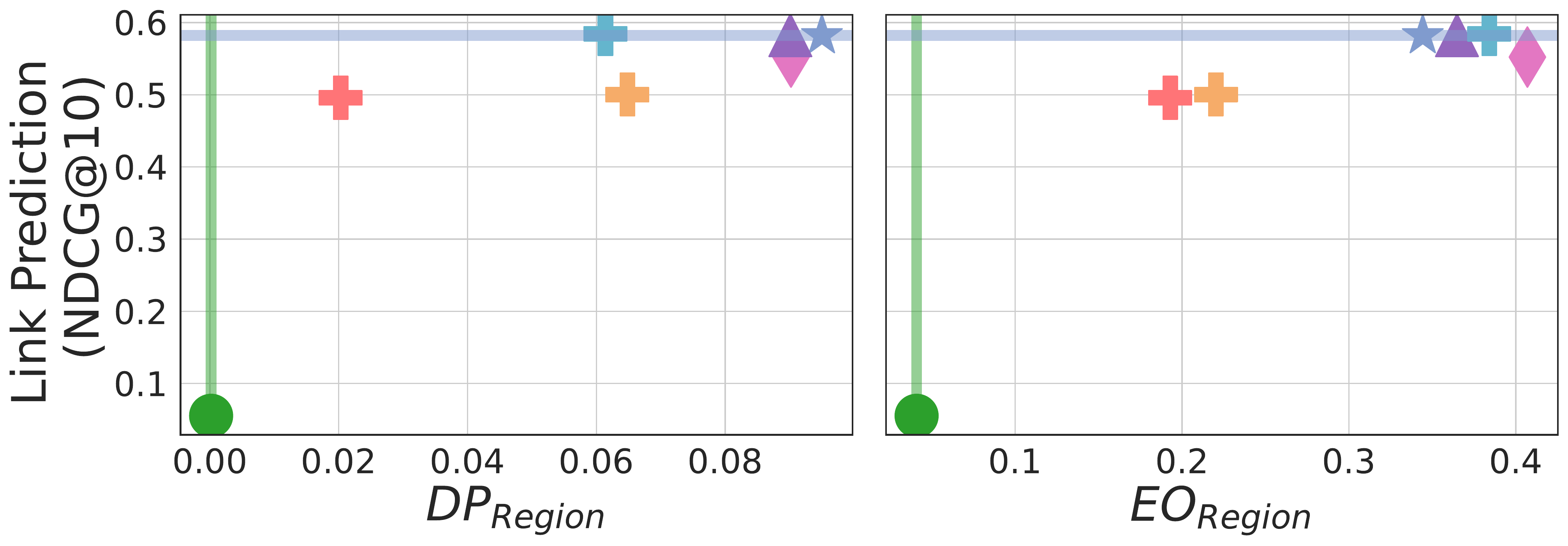}
 \end{subfigure}
 \hfill
 \begin{subfigure}[b]{0.88\linewidth}
 \centering
 \includegraphics[width=\linewidth]{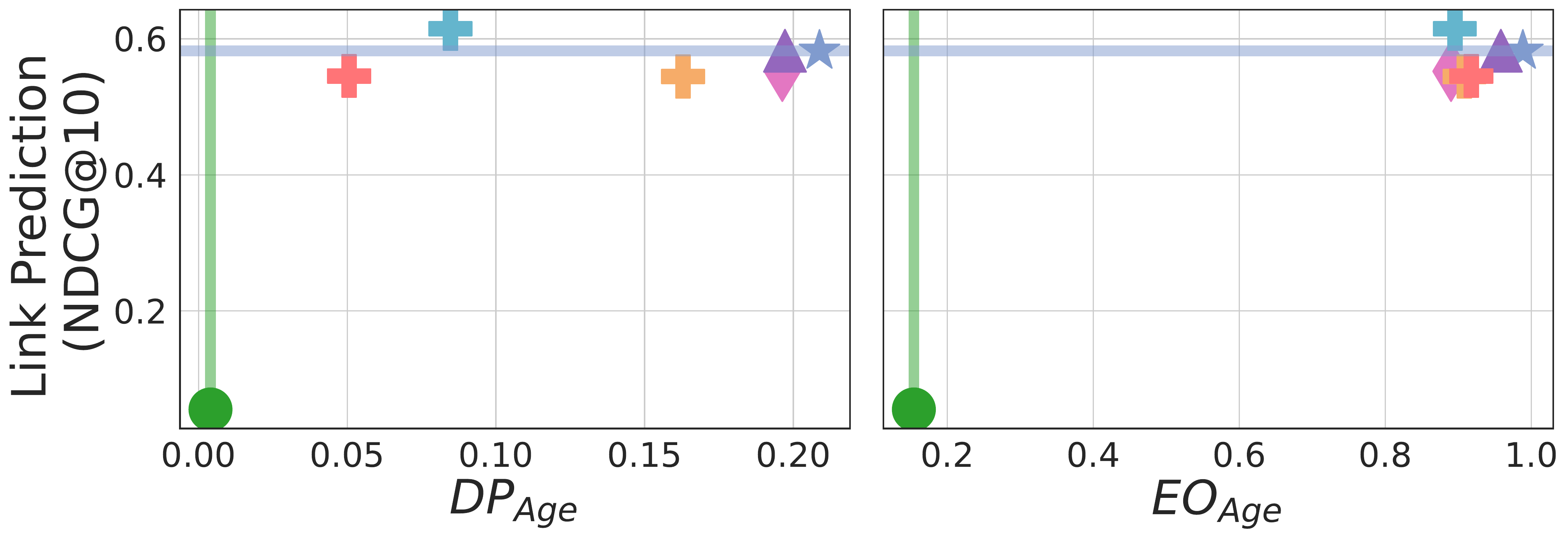}
 \end{subfigure}
 \hfill
 \caption{Fairness metrics evaluated on link prediction task on Pokec-n with node2vec as the embedding model.}
 \label{fig:fairness-pokecn}
 \vspace{-3mm}
\end{figure}

\subsection{High-Level Fairness from Embeddings}
\label{sec:fairness}
We study whether the debiased embeddings can lead to fairness in downstream tasks. We adopt two popular metrics---\emph{demographic parity} (DP) and \emph{equalized opportunity} (EO) to evaluate the fairness of link prediction results from the embeddings. DP requires that the predictions are independent from sensitive attributes, measured by the maximum difference of prediction rates between different combinations of  sensitive attribute values. EO measures the independence between true positive rate (TPR) of predicted edges and sensitive attributes. It is defined by the maximum difference of TPRs between different sensitive attribute value combinations. For both DP and EO, lower values suggest better fairness. We use the exact formulation of DP and EO in \cite{buyl2020debayes} and use the sigmoid function to convert the edge score for a pair of nodes to a probability. 

We show the results on fairness vs., utility in \Cref{fig:fairness-pokecn}, which are evaluated on each of the three sensitive attributes in Pokec-n with node2vec as the embedding model. In each plot, x-axis is the DP or EO and y-axis is the NDCG@10 on link prediction. Similar to \Cref{fig:tradeoff}, the ideal debiasing methods should locate at the upper left corner. Except for EO on the \emph{age} attribute where all methods performs similarly, \UGE methods can achieve significantly better fairness than the baselines on both DP and EO, while maintaining competitive performance on link prediction. \UGE-C can achieve the most fair predictions. This study shows \UGE's ability to achieve fairness in downstream tasks by effectively eliminating bias in the learned node embeddings.



\subsection{Unbiasedness-Utility Tradeoff in \UGE}
\label{sec:reg-weights}
Last but not least, we study the unbiasedness-utility trade-off in \UGE-C by tuning the weight on regularization. 
Although \UGE-W itself can already achieve promising debiasing effect, we expect that the added regularization from \UGE-R can complement it for a better trade-off. 
In particular, we tune the regularization weights in both CFC and \UGE-C and plot Micro-F1 (x-axis) vs. NDCG@10 (y-axis) from the resulting embeddings in \Cref{fig:bias-utility}.
Weight values are marked on each point and also listed in \Cref{appendix:settings}. The results are obtained on Pokec-z with GAT as the embedding model and the two figures correspond to debiasing \emph{gender} and \emph{region}, respectively. With the same extent of bias measured by Micro-F1, embeddings from \UGE-C have a much higher utility as indicated by the vertical grids. 
On the other hand, embeddings from \UGE-C have much less bias when the utility is the same as CFC, as indicated by horizontal grids. 
This experiment proves a better trade-off achieved in \UGE-C, which is consistent with our designs on \UGE-W and \UGE-R. \UGE-W learns from a bias-free graph without any constraints, and it is \emph{sufficient} to achieve unbiasedness without hurting the utility of the embeddings. \UGE-R constrains the embeddings to have the properties of those learned from a bias-free graph, which is \emph{necessary} for the embeddings to be unbiased. 

\begin{figure}[t]
 \centering
 \begin{subfigure}[b]{\linewidth}
 \centering
 \includegraphics[width=\linewidth]{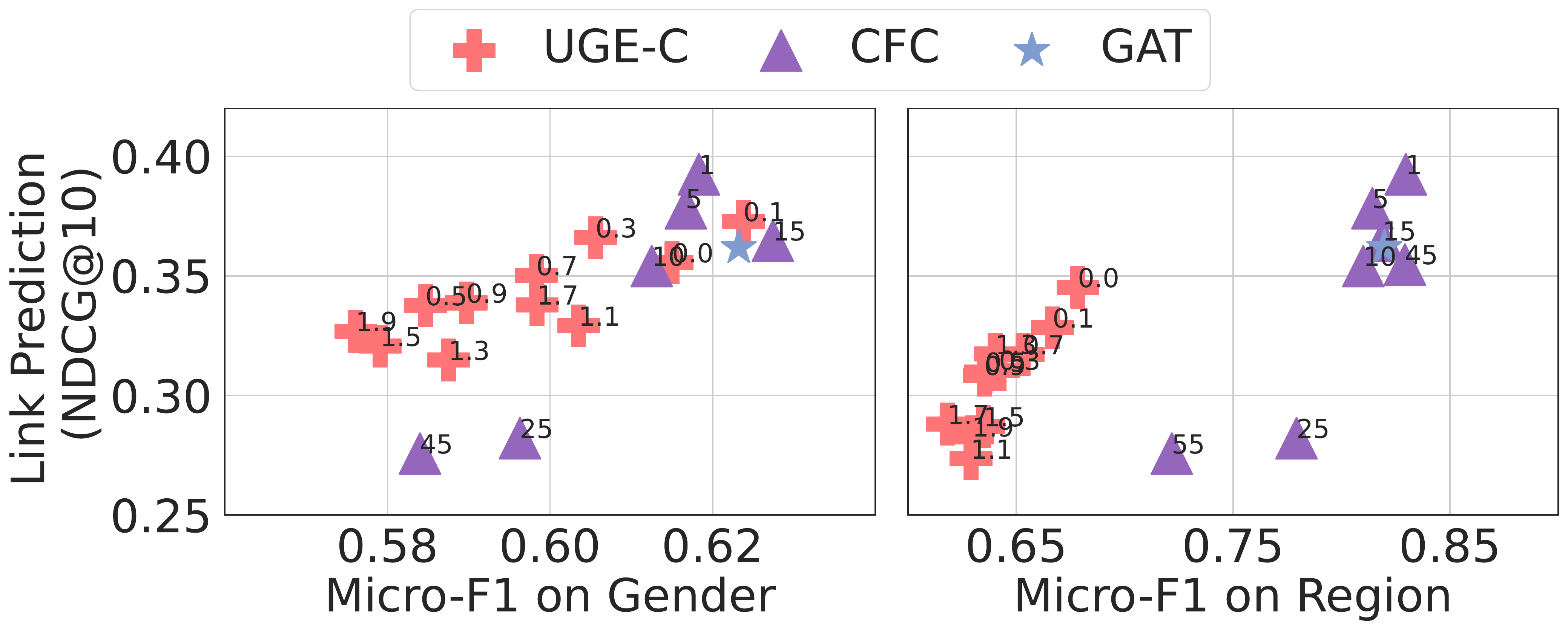}
 \end{subfigure}
 \caption{Trade-off comparison between CFC and \UGE-C on Pokec-z with GAT as the embedding model.}
 \label{fig:bias-utility}
 \vspace{-1em}
\end{figure}
\section{Conclusion}
\label{sec:conclusion}
We propose a principled new way for learning unbiased node embeddings from graphs biased by sensitive attributes. The idea is to infer a bias-free graph where the influence from sensitive attributes is removed, and then learn the node embeddings from it. This new perspective motivates our design of \UGE-W, \UGE-R and their combined methods \UGE-C. Extensive experiment results demonstrated strong debiasing effect from \UGE as well as better unbiasedness-utility trade-offs in downstream applications.

We expect the principle of \UGE can inspire better future designs for learning unbiased node embeddings from bias-free graphs. For example, instead of modeling the generation process and perform debiasing statistically, we can directly generate one or multiple bias-free graphs from the underlying generative graph model, and perform graph embedding on them. The regularization \UGE-R can be refined with better moment matching mechanism than minimizing the $l_2$ distance. The weights in \UGE-W can be modeled and learned for better debiasing effects. Besides, it is possible and promising to directly design unbiased GNN models that directly aggregate edges based on the inferred bias-free graph.

\begin{acks}
This work is supported by the National Science Foundation under grant IIS-1553568, IIS-1718216, IIS-2007492, and IIS-2006844.
\end{acks}

\bibliographystyle{ACM-Reference-Format}
\bibliography{reference}

\clearpage
\appendix
\section{Experimental Settings}
\label{appendix:settings}
Here we introduce more details about the experiment setup and model configurations for reproducibility.  

For GCN-type models (GCN, GAT, SGC), we use two convolutional layers with dimension $d_1=64$ and $d_2=16$.
For node2vec, we set walk length to 1 which turns a general skip-gram loss to objective of the link prediction task.
All the deep learning models are trained via Adam optimizer with step size $0.01$ for $800$ epochs, and we use a normalized weight decay $0.0005$ to prevent overfitting. 
Our proposed \UGE methods and the baseline CFC require a regularization weight to balance the task-specific objective and the debiasing effect. For CFC, we report the result with the regularization weight chosen from the set $\{1.0, 5.0, 10.0, 15.0, 25.0, 35.0, 45.0, 55.0, 65.0\}$, which finally is $\lambda=55.0$.
For \UGE, we test $\{0.1, 0.3, 0.5, 0.7, 0.9, 1.1,$ $1.3, 1.5, 1.7, 1.9\}$, and report the performance when $\lambda=0.5$. 
The regularization term in \eqref{eq:uge-r} is summed over all node pairs and can be costly to calculate. But empirically, $M$ group pairs sampled uniformly in each round of model update, where $M$ is around 10\% of the number of node groups, can already yield promising results. For evaluating the unbiasedness of the node embeddings, we use implementations from scikit-learn \cite{scikit-learn} for classifier training and evaluating Micro-F1.

\begin{figure}
 \centering
 \begin{subfigure}[b]{\linewidth}
 \centering
 \includegraphics[width=\textwidth]{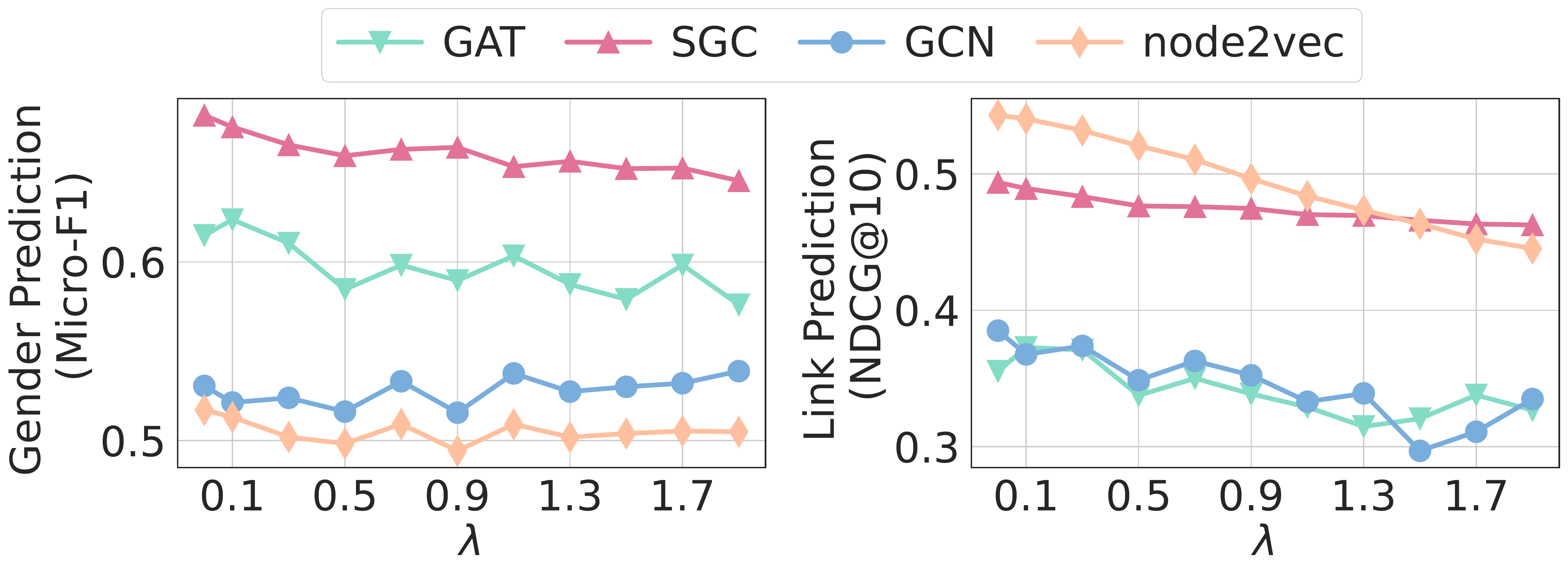}
 \end{subfigure}
 \begin{subfigure}[b]{\linewidth}
 \centering
 \includegraphics[width=\textwidth]{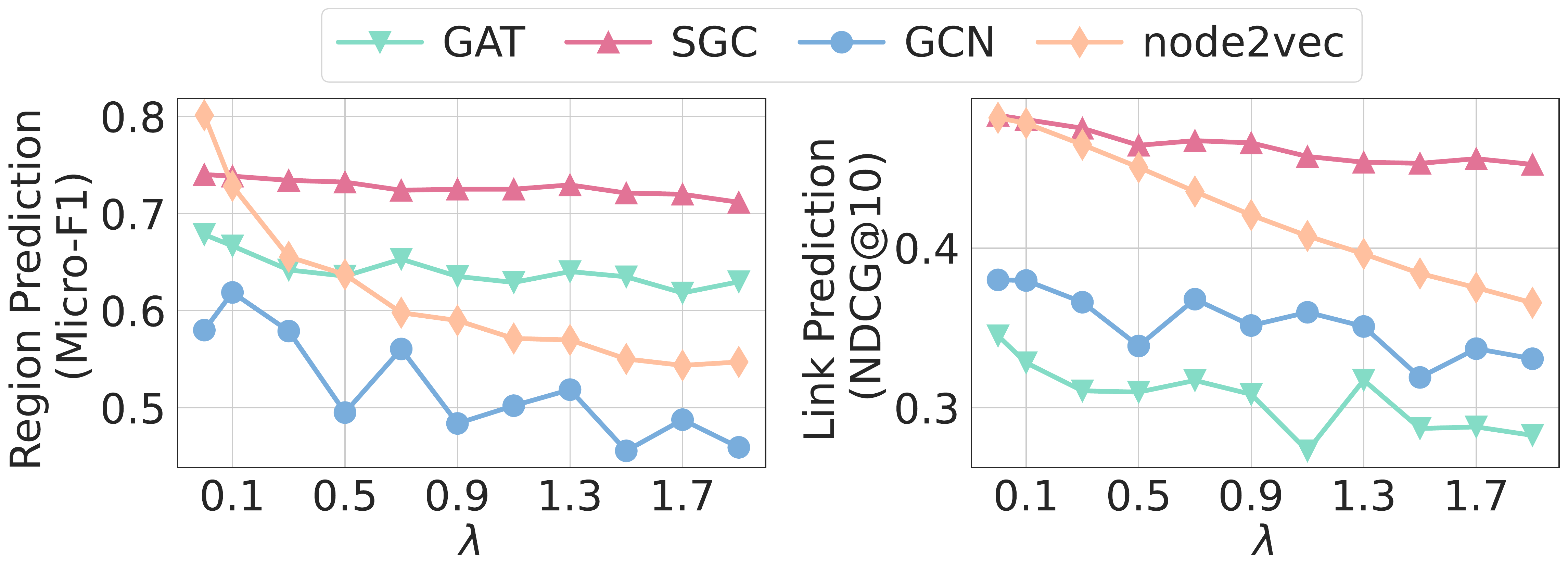}
 \end{subfigure}
 \begin{subfigure}[b]{\linewidth}
 \centering
 \includegraphics[width=\textwidth]{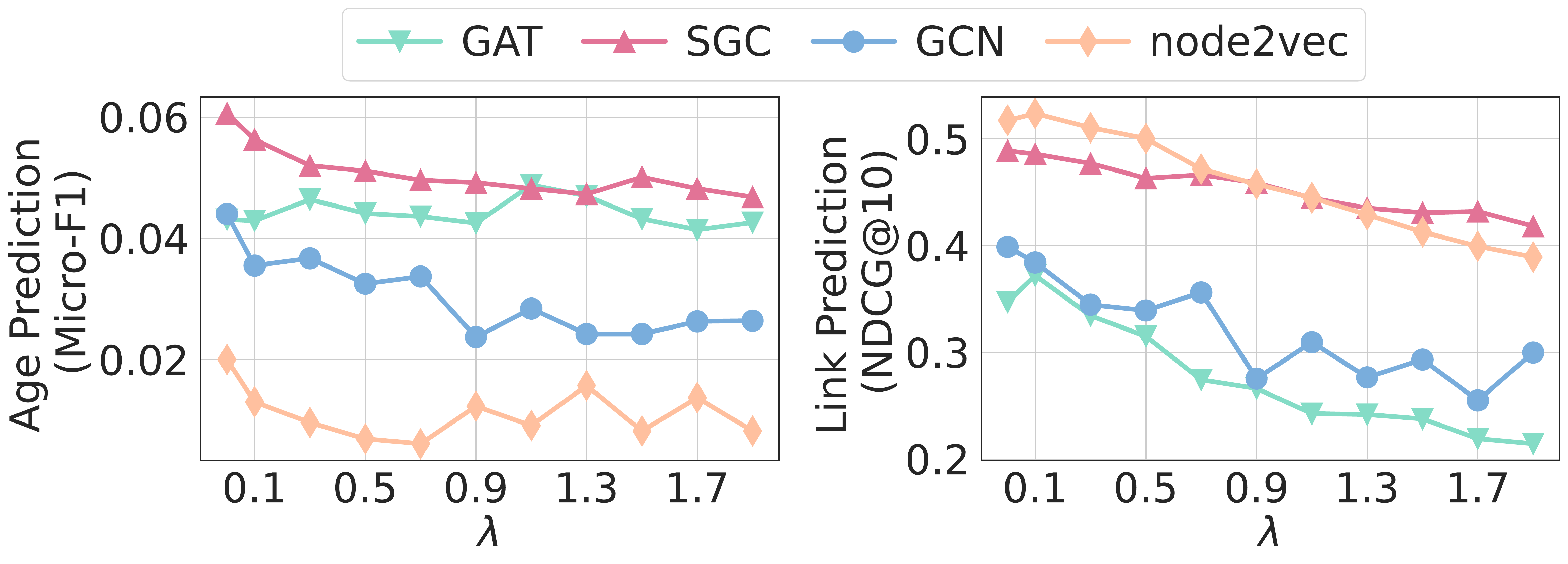}
 \end{subfigure}
 \caption{Unbiasedness and utility trade-off using different regularization weights on \UGE-C (x-axis). The left columns shows unbiasedness (attribute prediction), and the right columns shows utility (link prediction).}
 \label{apxfig:sensitivity}
\end{figure}


\begin{table*}[t]
\caption{The prediction performance of node embeddings learned on Pokec-z using four graph neural networks as embedding models. In each row, we use bold to mark the best debiasedness on attribute prediction or utility on link prediction.}
\begin{tabular}{ccccccccccccccc}
\toprule
& \multicolumn{1}{c}{} & &  & \multicolumn{5}{c}{Debiasing Method} & \multicolumn{1}{c}{}  & \multicolumn{1}{c}{} & \multicolumn{1}{c}{} & \multicolumn{1}{c}{} & \multicolumn{1}{c}{} & \multicolumn{1}{c}{} \\

\cmidrule(lr){5-9}
    \multirow{-2}{*}{Dataset}  & \multicolumn{1}{c}{\multirow{-2}{*}{Embedding Model}} & \multirow{-2}{*}{Prediction Target} & \multirow{-2}{*}{No Debiasing} & Fairwalk & CFC    & UGE-W  & UGE-R  & UGE-C  & \multicolumn{1}{c}{\multirow{-2}{*}{Random}} &  & &   &  &   \\
\midrule
    & \multicolumn{1}{c}{}  & Gender (Micro-F1)  & 0.6232  & 0.6135   & 0.5840 & 0.6150 & 0.6094 & \textbf{0.5747} & 0.4921  & \multicolumn{1}{c}{} & \multicolumn{1}{c}{} & \multicolumn{1}{c}{} & \multicolumn{1}{c}{} & \multicolumn{1}{c}{} \\
    & \multicolumn{1}{c}{}  & Link (NDCG@10) & 0.3618  & 0.3280   & 0.2757 & \textbf{0.3554} & 0.3422 & 0.3376 & 0.0570 & \multicolumn{1}{c}{} & \multicolumn{1}{c}{} & \multicolumn{1}{c}{} & \multicolumn{1}{c}{} & \multicolumn{1}{c}{} \\
\cmidrule(lr){3-10}
    & \multicolumn{1}{c}{}  & Region (Micro-F1)  & 0.8197& 0.8080   & 0.7217 & 0.6784 & 0.7660 & \textbf{0.6356} & 0.4966 & \multicolumn{1}{c}{} & \multicolumn{1}{c}{} & \multicolumn{1}{c}{} & \multicolumn{1}{c}{} & \multicolumn{1}{c}{} \\
    & \multicolumn{1}{c}{}  & Link (NDCG@10) & 0.3618  & 0.3287   & 0.2757 & 0.3451 & \textbf{0.3547} & 0.3098 & 0.0570& \multicolumn{1}{c}{} & \multicolumn{1}{c}{} & \multicolumn{1}{c}{} & \multicolumn{1}{c}{} & \multicolumn{1}{c}{} \\
 \cmidrule(lr){3-10}
  & \multicolumn{1}{c}{} & Age (Micro-F1)& 0.0526 & 0.0522   & 0.0498 & 0.0431 & 0.0545 & \textbf{0.0429} & 0.0007  & \multicolumn{1}{c}{} & \multicolumn{1}{c}{} & \multicolumn{1}{c}{} & \multicolumn{1}{c}{} & \multicolumn{1}{c}{} \\
  & \multicolumn{1}{c}{\multirow{-6}{*}{GAT}} & \multicolumn{1}{l}{Link (NDCG@10)}  & 0.3618& 0.3122   & 0.2757 & 0.3471 & 0.3205 & \textbf{0.3718} & 0.0570  & &  &  &  &  \\
 \cmidrule(lr){2-10}
    &  & Gender (Micro-F1) & 0.6766  & 0.6631   & \textbf{0.6520} & 0.6822 & 0.6531 & 0.6596 & 0.4921   &  & & &  & \\
    &  & Link (NDCG@10)    & 0.4975   & 0.4461   & 0.4011 & \textbf{0.4938} & 0.4850 & 0.4765 & 0.0570    & &      &    &   &    \\
 \cmidrule(lr){3-10}
 &    & Region (Micro-F1) & 0.7806                & 0.7820   & \textbf{0.7150} & 0.7402 & 0.7680 & 0.7323 & 0.4966   &   &     &     &                      &     \\
 &   & Link (NDCG@10) & 0.4975 & 0.4460   & 0.4011 & \textbf{0.4832} & 0.4799 & 0.4644 & 0.0570                                       &                      &                      &                      &                      &                      \\
  \cmidrule(lr){3-10}
 &  & Age (Micro-F1)                      & 0.0621                         & 0.0662   & 0.0654 & 0.0606 & 0.0529 & \textbf{0.0510} & 0.0007                                       &                      &                      &                      &                      &                      \\
 & \multirow{-6}{*}{SGC}                                 & \multicolumn{1}{l}{Link (NDCG@10)}  & 0.4975                         & 0.4461   & 0.4011 & \textbf{0.4889} & 0.4694 & 0.4630 & 0.0570                                       &                      &                      &                      &                      &                      \\
  \cmidrule(lr){2-10}
 &                                                       & Gender (Micro-F1)                   & 0.5532                         & 0.5589   & 0.5493 & 0.5306 & 0.5301 & \textbf{0.5162} & 0.4921                                       &                      &                      &                      &                      &                      \\
 &                                                       & Link (NDCG@10)                      & 0.3865                         & 0.2807   & 0.3836 & \textbf{0.3851} & 0.3727 & 0.3488 & 0.0570                                       &                      &                      &                      &                      &                      \\
  \cmidrule(lr){3-10}
 &                                                       & Region (Micro-F1)                   & 0.7445                         & 0.7616   & 0.7693 & 0.5800 & 0.6105 & \textbf{0.4951} & 0.4966                                       &                      &                      &                      &                      &                      \\
 &                                                       & Link (NDCG@10)                      & 0.3865                         & 0.2807   & \textbf{0.3836} & 0.3801 & 0.3360 & 0.3386 & 0.0570                                       &                      &                      &                      &                      &                      \\
  \cmidrule(lr){3-10}
 &                                                       & Age (Micro-F1)                      & 0.0425                         & 0.0416   & 0.0391 & 0.0439 & 0.0409 & \textbf{0.0324} & 0.0007                                       &                      &                      &                      &                      &                      \\
 & \multirow{-6}{*}{GCN}                                 & \multicolumn{1}{l}{Link (NDCG@10)}  & 0.3865                         & 0.2807   & 0.3836 & \textbf{0.3987} & 0.3550 & 0.3391 & 0.0570                                       &                      &                      &                      &                      &                      \\
  \cmidrule(lr){2-10}
 &                                                       & Gender (Micro-F1)                   & 0.5248                         & 0.5347   & 0.5137 & 0.5171 & \textbf{0.4949} & 0.4982 & 0.4921                                       &                      &                      &                      &                      &                      \\  &                                                       & Link (NDCG@10)                      & 0.5491                         & 0.5120   & \textbf{0.5496} & 0.5430 & 0.5463 & 0.5206 & 0.0570                                       &                      &                      &                      &                      &                      \\
  \cmidrule(lr){3-10}
 &                                                       & Region (Micro-F1)                   & 0.8423                         & 0.8462   & 0.8423 & 0.8012 & 0.6490 & \textbf{0.6372} & 0.4966                                       &                      &                      &                      &                      &                      \\
 &                                                       & Link (NDCG@10)                      & 0.5491                         & 0.5120   & \textbf{0.5496} & 0.4816 & 0.5354 & 0.4506 & 0.0570                                       &                      &                      &                      &                      &                      \\
  \cmidrule(lr){3-10}
 &                                                       & Age (Micro-F1)                      & 0.0365                         & 0.0404   & 0.0365 & 0.0200 & 0.0122 & \textbf{0.0068} & 0.0007                                       &                      &                      &                      &                      &                      \\
\multirow{-24}{*}{Pokec-z} & \multirow{-6}{*}{node2vec}                            & \multicolumn{1}{l}{Link (NDCG@10)}  & 0.5491                         & 0.5120   & \textbf{0.5496} & 0.5173 & 0.5439 & 0.5002 & 0.0570                                       &                      &                      &                      &                      &  
\\
\bottomrule
\label{apx:pokecz}
\end{tabular}
\end{table*}

\section{Results}
\label{appendix:results}
In \Cref{appendix:more-results}, we include additional experiment results to report the trade-off between unbiasedness and utility on the complete set of embedding models on Pokec-z. 
In \Cref{apx:ablation}, we show a complete comparison among our proposed instances of unbiased graph embedding \UGE-W, \UGE-R and \UGE-C.
In \Cref{apx:sensitivity}, we investigate the influence of the regularization weight on the complete set of embedding models.

\subsection{Additional Analysis on Undebiasedness}
\label{appendix:more-results}
\Cref{apx:pokecz} summarizes the debiasing and utility performance of the proposed method and baselines when using four graph neural networks on Pokec-z.
Each line of attribute prediction result is followed by the corresponding performance on link prediction.
Generally, \UGE-W achieves the best link prediction performance and \UGE-R has better debiasing effect. Combining \UGE-W with \UGE-R produces \UGE-C with better trade-off.

\subsection{Ablation Study}
\label{apx:ablation}

\begin{figure}
 \centering
 \begin{subfigure}[b]{\linewidth}
 \centering
 \includegraphics[width=\textwidth]{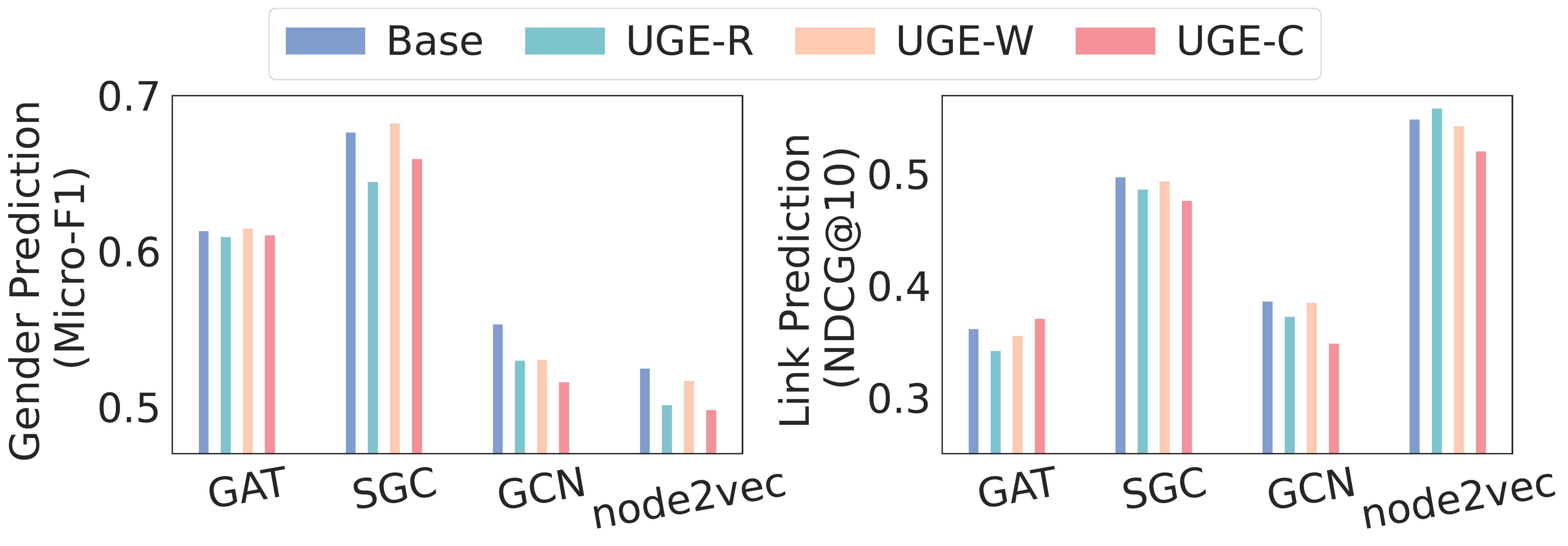}
 \end{subfigure}
 \begin{subfigure}[b]{\linewidth}
 \centering
 \includegraphics[width=\textwidth]{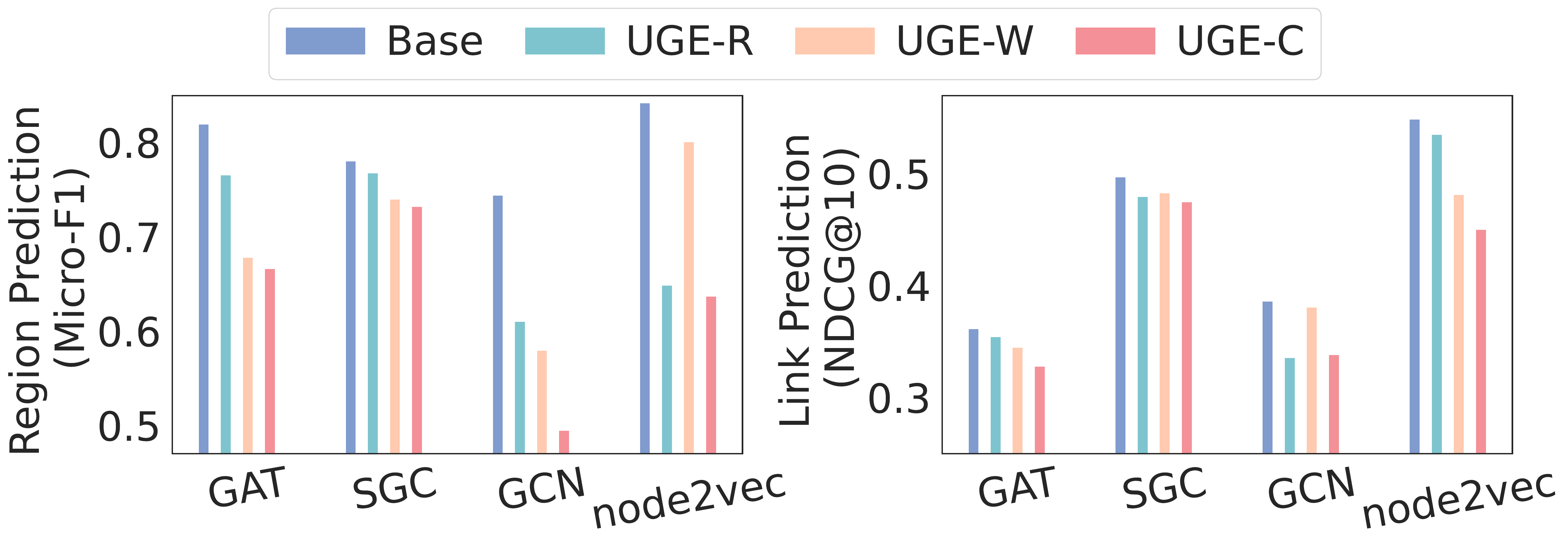}
 \end{subfigure}
 \begin{subfigure}[b]{\linewidth}
 \centering
 \includegraphics[width=\textwidth]{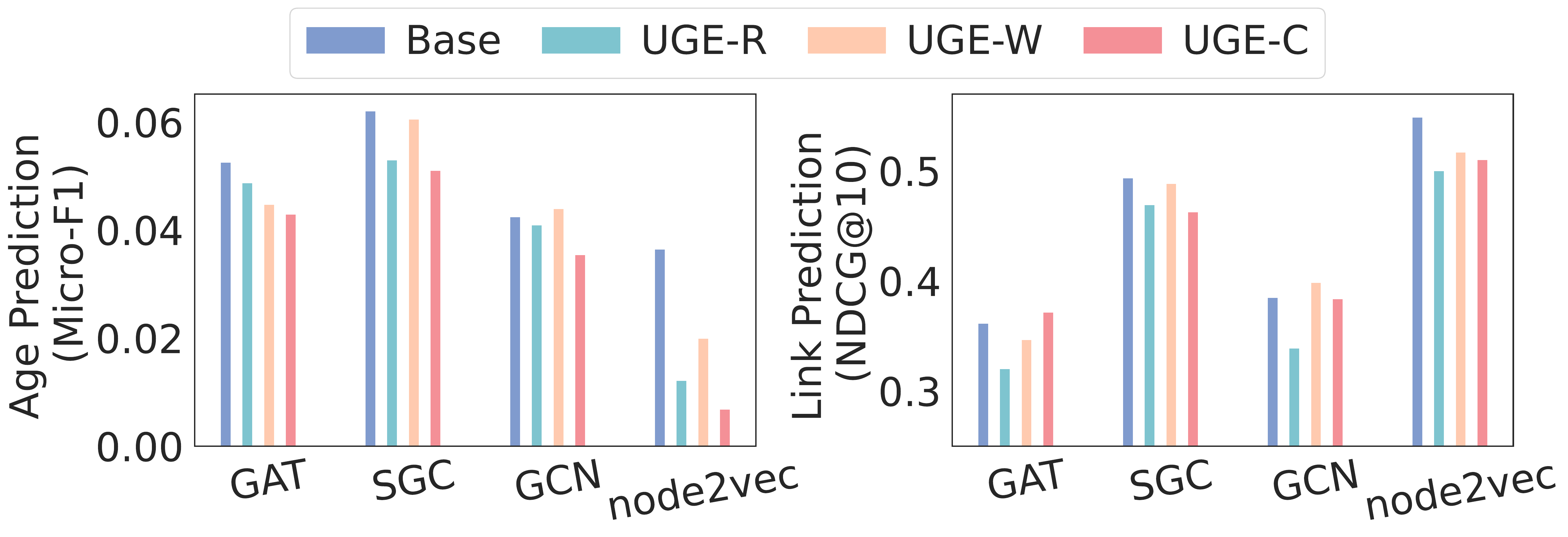}
 \end{subfigure}
 \caption{Comparison among our proposed models on different embedding models. The left columns shows the unbiasedness (attribute prediction) and the right columns shows the utility (link prediction).}
 \label{apxfig:ablation}
\end{figure}

\Cref{apxfig:ablation} presents the performance of three proposed model (\UGE-W, \UGE-R and \UGE-C) applied to four graph neural networks (GAT, SGC, GCN and node2vec). We can clearly observe that in most cases \UGE-R has better debiasing effect compared with \UGE-W, while \UGE-W can better maintain the utility for downstream link prediction task. \UGE-C as the combination of them indeed makes the best of the both designs.

\subsection{Unbiasedness-Utility Tradeoff in \UGE}
\label{apx:sensitivity}

In addition to \Cref{sec:reg-weights} where we only showed the effect of regularization weight on Pokec-z with GAT as the embedding model, we now include a complete analysis on unbiasedness and utility trade-off across embedding models in \Cref{apxfig:sensitivity}. It clearly shows a trade-off: as the weight increases, we obtain a stronger debiasing effect with a cost of the utility on link prediction.

\end{document}